\documentclass[10pt,twocolumn,letterpaper]{article}

\usepackage{cvpr}              % To produce the CAMERA-READY version
\usepackage{graphicx}
\usepackage{amsmath}
\usepackage{amssymb}
\usepackage{booktabs}
\usepackage[accsupp]{axessibility}

\usepackage[pagebackref,breaklinks,colorlinks]{hyperref}

\usepackage[capitalize]{cleveref}
\crefname{section}{Sec.}{Secs.}
\Crefname{section}{Section}{Sections}
\Crefname{table}{Table}{Tables}
\crefname{table}{Tab.}{Tabs.}

\usepackage{multirow}
\usepackage{booktabs}
\usepackage{makecell}
\usepackage{adjustbox}

\usepackage{color}

\usepackage[normalem]{ulem}

\definecolor{ben}{rgb}{0.9,0.,0.5}

\definecolor{pat}{rgb}{0.6,0.2,0.1}

\definecolor{GY}{rgb}{0.4,0.1,0.9}

\newcommand{\mycomment}[1]{}

\newlength\savewidth

\def\cvprPaperID{9363} % *** Enter the CVPR Paper ID here
\def\confName{CVPR}
\def\confYear{2023}

\begin{document}

\title{On the Importance of Accurate Geometry Data for Dense 3D Vision Tasks}

\author{
\hspace{-20pt}
HyunJun Jung$^{\ast 1}$,
Patrick Ruhkamp$^{\ast 1,2}$,
Guangyao Zhai$^{1}$,
Nikolas Brasch$^{1}$,
Yitong Li$^{1}$,\\
Yannick Verdie$^{1,3}$,
Jifei Song$^{3}$,
Yiren Zhou$^{3}$,
Anil Armagan$^{3}$,
Slobodan Ilic$^{1,4}$,\\
Ales Leonardis$^{3}$,
Nassir Navab$^{1}$,
Benjamin Busam$^{1,2}$
\\
\\
\small
$^1$ Technical University of Munich,
$^2$ 3Dwe.ai,
$^3$ Huawei Noah's Ark Lab,
$^4$ Siemens AG,
$^*$ Equal Contribution
\\ \footnotesize{\fontfamily{qcr}\selectfont
hyunjun.jung@tum.de, p.ruhkamp@tum.de, guangyao.zhai@tum.de, b.busam@tum.de
}
}

\maketitle

%%%%%%%%% ABSTRACT
\begin{abstract}

Learning-based methods to solve dense 3D vision problems typically train on 3D sensor data. The respectively used principle of measuring distances provides advantages and drawbacks. These are typically not compared nor discussed in the literature due to a lack of multi-modal datasets. Texture-less regions are problematic for structure from motion and stereo, reflective material poses issues for active sensing, and distances for translucent objects are intricate to measure with existing hardware. Training on inaccurate or corrupt data induces model bias and hampers generalisation capabilities. These effects remain unnoticed if the sensor measurement is considered as ground truth during the evaluation. This paper investigates the effect of sensor errors for the dense 3D vision tasks of depth estimation and reconstruction. We rigorously show the significant impact of sensor characteristics on the learned predictions and notice generalisation issues arising from various technologies in everyday household environments. For evaluation, we introduce a carefully designed dataset\footnote{dataset available at https://github.com/Junggy/HAMMER-dataset} comprising measurements from commodity sensors, namely D-ToF, I-ToF, passive/active stereo, and monocular RGB+P. Our study quantifies the considerable sensor noise impact and paves the way to improved dense vision estimates and targeted data fusion.

\end{abstract}

%%%%%%%%% BODY TEXT

\section{Introduction}
\label{sec:introduction}
Our world is 3D. Distance measurements are essential for machines to understand and interact with our environment spatially. 
Autonomous vehicles~\cite{geiger2012we,ruhkamp2021attention,kong2020semantic,su2023opa} need this information to drive safely, robot vision requires distance information to manipulate objects~\cite{fang2020graspnet,wang2021demograsp,da2dataset,zhai2022monograspnet}, and AR realism benefits from spatial understanding~\cite{kopf2021robust,busam2019sterefo}.

\begin{figure}[t]
\begin{center}
\includegraphics[width=1.00\linewidth]{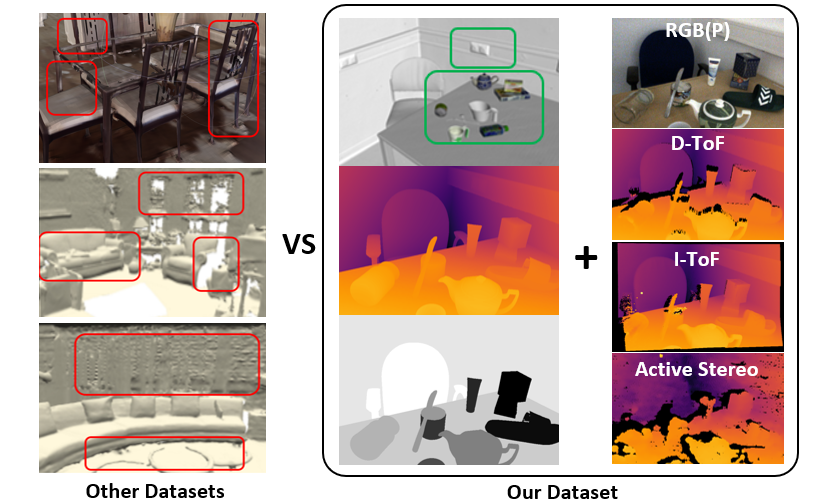}
\vspace{-0.9cm}
\end{center}
  \caption{Other datasets for dense 3D vision tasks reconstruct the scene as a whole in one pass~\cite{replica19arxiv,dai2017scannet,Matterport3D}, resulting in low quality and accuracy (cf. red boxes). On the contrary, our dataset scans the background and every object in the scene separately a priori and annotates them as dense and high-quality 3D meshes. Together with precise camera extrinsics from robotic forward-kinematics, this enables a fully dense rendered depth as accurate pixel-wise ground truth with multimodal sensor data, such as RGB with polarization, D-ToF, I-ToF and Active Stereo. Hence, it allows quantifying different downstream 3D vision tasks such as monocular depth estimation, novel view synthesis, or 6D object pose estimation.}
\label{fig:teaser}
\vspace{-0.2cm}
\end{figure}

A variety of sensor modalities and depth prediction pipelines exist.
The computer vision community thereby benefits from a wide diversity of publicly available datasets~\cite{scharstein2002taxonomy,geiger2012we,silberman2012indoor,sturm2012benchmark,xiao2013sun3d,PhoCal,CroMo}, which allow for evaluation of depth estimation pipelines.
Depending on the setup, different sensors are chosen to provide ground truth (GT) depth maps, all of which have their respective advantages and drawbacks determined by their individual principle of distance reasoning.
Pipelines are usually trained on the data without questioning the nature of the depth sensor used for supervision and do not reflect areas of high or low confidence of the GT.

Popular \textbf{passive sensor} setups include multi-view stereo cameras where the known or calibrated spatial relationship between them is used for depth reasoning~\cite{scharstein2002taxonomy}. 
Corresponding image parts or patches are photometrically or structurally associated, and geometry allows to triangulate points within an overlapping field of view.
Such photometric cues are not reliable in low-textured areas and with little ambient light where \textbf{active sensing} can be beneficial~\cite{silberman2012indoor,sturm2012benchmark}.
Active stereo can be used to artificially create texture cues in low-textured areas and photon-pulses with a given sampling rate are used in Time-of-Flight (ToF) setups either directly (D-ToF) or indirectly (I-ToF)~\cite{Guo_2018_ECCV}.
With the speed of light, one can measure the distance of objects from the return time of the light pulse, but unwanted multi-reflection artifacts also arise.
Reflective and translucent materials are measured at incorrect far distances, and multiple light bounces distort measurements in corners and edges. 
While ToF signals can still be aggregated for dense depth maps, a similar setup is used with LiDAR sensors which sparsely measure the distance using coordinated rays that bounce from objects in the surrounding.
The latter provides ground truth, for instance, for the popular outdoor driving benchmark KITTI~\cite{geiger2012we}.
While LiDAR sensing can be costly, radar~\cite{gasperini2021r4dyn}  provides an even sparser but more affordable alternative.
\textbf{Multiple modalities} can also be fused to enhance distance estimates.
A common issue, however, is the inherent problem of warping onto a common reference frame which requires the information about depth itself~\cite{lopez2020project,jung2021wild}.
While multi-modal setups have been used to enhance further monocular depth estimation using self-supervision from stereo and temporal cues~\cite{monodepth2,CroMo}, its performance analysis is mainly limited to average errors and restricted by the individual sensor used.
An unconstrained analysis of depth in terms of RMSE compared against a GT sensor only shows part of the picture as different sensing modalities may suffer from drawbacks.

Where are the drawbacks of current depth-sensing modalities - and how does this impact pipelines trained with this (potentially partly erroneous) data?
Can self- or semi-supervision overcome some of the limitations posed currently?
To objectively investigate these questions, we provide multi modal sensor data as well as highly accurate annotated depth so that one can analyse the deterioration of popular monocular depth estimation and 3D reconstruction methods (see Fig.~\ref{fig:teaser}) on areas of different photometric complexity and with varying structural and material properties while changing the sensor modality used for training.
To quantify the impact of sensor characteristics, we build a unique camera rig comprising a set of the most popular indoor depth sensors and acquire synchronised captures with highly accurate ground truth data using 3D scanners and aligned renderings.
To this end, our main contributions can be summarized as follows:
\begin{enumerate}
    \item We question the measurement quality from commodity \textbf{depth sensor} modalities and analyse their \textbf{impact} as supervision signals for the dense 3D vision tasks of depth estimation and reconstruction.
    \item We investigate performance on texture-varying material as well as \textbf{photometrically challenging} reflective, translucent and transparent \textbf{areas} where \textbf{learning methods} systematically \textbf{reproduce sensor errors}.
    \item To objectively assess and quantify different data sources, we contribute an \textbf{indoor dataset} comprising an unprecedented combination of \textbf{multi-modal sensors}, namely I-ToF, D-ToF, monocular RGB+P, monochrome stereo, and active light stereo together with highly accurate ground truth. 
\end{enumerate}
\section{Related Work}
\label{sec:relatedwork}
\subsection{Geometry from X}
A variety of sensor modalities have been used to obtain depth maps.
Typical datasets comprise one ground truth sensor used for all acquisitions, which is assumed to give accurate enough data to validate the models:

\smallskip
\noindent
\textit{Stereo Vision.}
In the stereo literature, early approaches~\cite{scharstein2002taxonomy} use a pair of passive cameras and restrict scenes to piecewise planar objects for triangulation.
Complex setups with an industrial robot and structured light can yield ground truth depth for stereo images~\cite{aanaes2016IJCV}. Robots have also been used to annotate keypoints on transparent household objects~\cite{liu2020keypose}.
As these methods are incapable of retrieving reliable depth in textureless areas where stereo matching fails, active sensors are used to project patterns onto the scenes to artificially create structures.
The availability of active stereo sensors makes it also possible to acquire real indoor environments~\cite{silberman2012indoor} where depth data at missing pixels is inpainted.
Structure from motion (SfM) is used to generate the depth maps of Sun3D~\cite{xiao2013sun3d} where a moving camera acquires the scenes and data is fused ex post. 
A temporally tracked handheld active sensor is further used for depth mapping for SLAM evaluation in the pioneering dataset of Sturm et al.~\cite{sturm2012benchmark}.
While advancing the field, its depth maps are limited to the active IR-pattern used by its RGB-D sensor.

\begin{figure*}[!t]
 \centering
    \includegraphics[width=\linewidth]{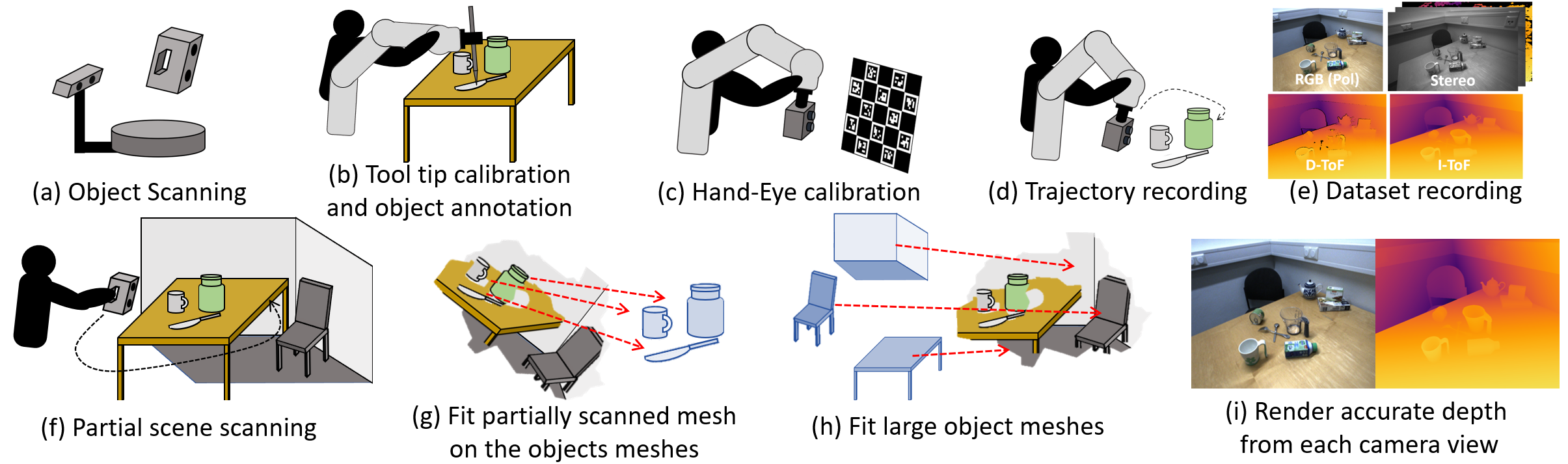}
    \vspace{-5mm}
    \caption{\textbf{Scanning Process Overview.} To extract highly accurate geometry, we design a multi-stage acquisition process. At first, 3D models are extracted with structured light 3D scanners (a). Scene objects (b) and mounted sensor rig (b) are calibrated towards a robot for accurate camera pose retrieval~\cite{PhoCal}. A motion trajectory is recorded in gravity compensation mode (d) and repeated to record synchronized images of all involved sensors (e). A partial digital twin of the 3D scene (f) is aligned to small (g) and larger (h) objects to retrieve an entire in silico replica of the scene which can be rendered from the camera views of each sensor used (i) which results in highly accurate dense depth maps that enable investigations of individual sensor components.}
    \label{fig:annotation_overview}
\end{figure*}

\smallskip
\noindent
\textit{Time-of-Flight Sensors.}
Further advances in active depth sensing emphasize ToF more. 
Initial investigations focus on simulated data~\cite{Guo_2018_ECCV} and controlled environments with little ambient noise~\cite{son2016learning}. 
The broader availability of ToF sensors in commercial products (e.g. Microsoft Kinect series) and modern smartphones (e.g. I-ToF of Huawei P30 Pro, D-ToF in Apple iPhone 12) creates a line of research around curing the most common sensor errors. These are multi-path interference (MPI), motion artefacts and a high level of sparsity and shot noise~\cite{jung2021wild}. Aside of classical active and passive stereo, we therefore also include D-ToF and I-ToF modalities in all our experiments.

\smallskip
\noindent
\textit{Polarimetric Cues.}
Other properties of light are used to indirectly retrieve scene surface properties in the form of normals for which the amount of linearly polarized light and its polarization direction provide information, especially for highly reflective and transparent objects~\cite{kalra2020deep,gao2021polarimetric}. 
Initial investigations for shape from polarization mainly analyse controlled setups~\cite{garcia2015surface,atkinson2006recovery,yu2017shape,smith2018height}. More recent approaches investigate also sensor fusion methods~\cite{kadambi2017depth} even in challenging scenes with strong ambient light~\cite{CroMo}. 
We consequently also acquire RGB+P data for all scenes.

\smallskip
\noindent
\textit{Synthetic Renderings.}
In order to produce pixel-perfect ground truth, some scholars render synthetic scenes~\cite{mayer2016large}. 
While this produces the best possible depth maps, the scenes are artificially created and lack realism, causing pipelines trained on Sintel~\cite{Butler:ECCV:2012} or SceneFlow~\cite{mayer2016large} to suffer from a synthetic-to-real domain gap. 
In contrast, we follow a hybrid approach and leverage pixel-perfect synthetic data from modern 3D engines to adjust highly accurate 3D models to real captures.

\begin{figure*}[!t]
 \centering
    \includegraphics[width=\linewidth]{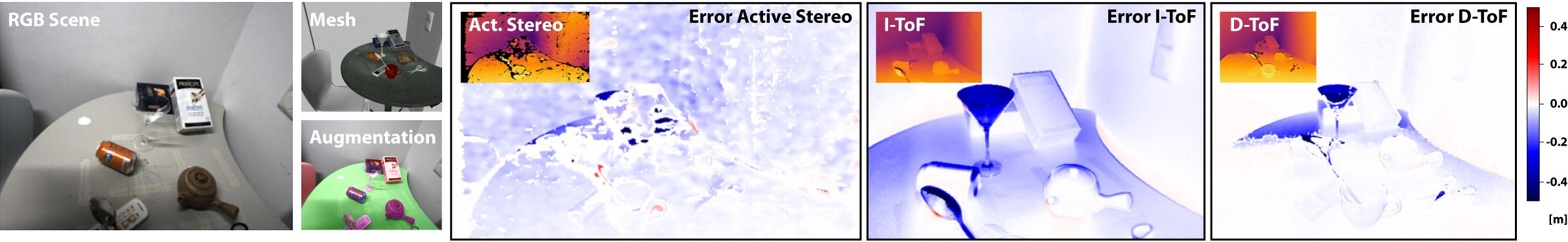}
    \vspace{-7mm}
    \caption{\textbf{Data Quality.} A full 3D reconstruction of the RGB scene (left) allows to render highly accurate depth maps from arbitrary views. These serve as GT to study sensor errors of various depth sensors for different scene structures (right). E.g., due to the measurement principle, the translucent glass becomes invisible for the ToF sensors.}
    \label{fig:dataset_quality}
\end{figure*}

\subsection{Monocular Depth Estimation}
Depth estimation from a single image is inherently ill-posed. 
Deep learning has enabled this task for real scenes. 

\smallskip
\noindent
\textit{Supervised Training.}
Networks can learn to predict depth with supervised training. Eigen et al.~\cite{eigen2014depth} designed the first monocular depth estimation network by learning to predict coarse depth maps, which are then refined by a second network. Laina et al.~\cite{laina2016deeper} improved the latter model by using only convolutional layers in a single CNN. 
The required ground truth often limits these methods to outdoor scenarios~\cite{Geiger_2013}. 
A way of bypassing this is to use synthetic data~\cite{mayer2018makes}. 
Narrowing down the resulting domain gap can be realized~\cite{Guo_2018_ECCV}.  MiDaS~\cite{Ranftl2022} generalizes better to unknown scenes by mixing data from 3D movies. 
To predict high-resolution depth, most methods use multi-scale features or post processing~\cite{Miangoleh2021Boosting,Wei2021CVPR} which complicates learning. 
If not trained on a massive set of data, these methods show limited generalization capabilities.

\smallskip
\noindent
\textit{Self-Supervision.}
Self-supervised monocular methods try to circumvent this issue. 
The first such methods~\cite{xie2016deep3d,garg2016unsupervised} propose to use stereo images to train a network for depth prediction. With it, the left image is warped into the right where photometric consistency serves as training signal. 
Monodepth~\cite{Godard_2017} added a left-right consistency loss to mutually leverage warping from one image into the other. 
Even though depth quality improves, it requires synchronized image pairs. 
Monocular training methods are developed that use only one camera where frames in a video are leveraged for the warping with simultaneously estimated poses between them. This task is more intricate, however, Monodepth2~\cite{monodepth2} reduces the accuracy gap between the stereo and monocular training by automasking and with a minimum reprojection loss. 
A large body of work further improves the task~\cite{yang2018lego,chen2019towards,spencer2020defeat,lee2021patch,ranftl2021vision,Ranftl2022} and investigates temporal consistency~\cite{luo2020consistent,watson2021temporal,ruhkamp2021attention}. 
To compare the effect of various supervision signals for monocular depth estimation, we utilized the ResNet backbone of the popular Monodepth2~\cite{monodepth2} together with its various training strategies.

\subsection{Reconstruction and Novel View Synthesis}
The 3D geometry of a scene can be reconstructed from 2D images and optionally their depth maps~\cite{kinectfusion}. Scenes are stored explicitly or implicitly. Typical explicit representation include point clouds or meshes~\cite{curless1996volumetric} while popular implicit representation are distance fields~\cite{zeng20163dmatch} which provide the scene as a level set of a given function, or neural fields where the scene is stored in the weights of a network~\cite{xie2022}.

\smallskip
\noindent
\textit{NeRFs.}
Due to their photorealism in novel view synthesis, recent advances around neural radiance fields (\textit{NeRF})~\cite{mildenhall2021nerf} experience severe attention. 
In this setup, one network is trained on a posed set of images to represent a scene. The method optimizes for the prediction of volume density and view-dependent emitted radiance within a volume. Integration along query rays allows to synthesize novel views of static and deformable~\cite{park2021nerfies} scenes.
Most noticeable recent advances extend the initial idea to unbounded scenes of higher quality with Mip-NeRF 360~\cite{barron2022mip} or factor the representation into low-rank components with TensoRF~\cite{chen2022tensorf} for faster and more efficient usage. Also robustness to pose estimates and calibration are proposed~\cite{lin2021barf,wang2021nerf}. 
While the initial training was computationally expensive, methods have been developed to improve inference and training. 
With spherical harmonics spaced in a voxel grid structure, Plenoxels~\cite{fridovich2022plenoxels} speed up processes even without a neural network and interpolation techniques~\cite{sun2022direct} accelerate training. 
Geometric priors such as sparse and dense depth maps can regularize convergence, improve quality and training time~\cite{deng2022depth,roessle2022dense}.
Besides recent works on methods themselves, \cite{Reizenstein_2021_ICCV} propose to leverage real world objects from crowd-sourced videos on a category level to construct a dataset to evaluate novel view synthesis and category-centric 3D reconstruction methods.

We make use of most recent NeRF advances and analyse the impact of sensor-specific depth priors in~\cite{roessle2022dense} for the task of implicit scene reconstruction.
To neglect the influence of pose estimates and produce highly accurate data, we leverage the robotic pose GT of our dataset. 
\section{Data Acquisition \& Sensor Modalities}
\label{sec:dataset}

% \begin{figure}[!b]
%  \centering
%     \includegraphics[width=\linewidth]{figures/hammer_other_datasets_comp.pdf}
%     \caption{\textbf{Issues of Other Datasets.} Multiple sources of error, e.g. missing glass surfaces (table in left and in right scenes), noisy background, missing data, corrupted thin objects, holes, etc. exist in Replica~\cite{replica19arxiv}, ScanNet~\cite{dai2017scannet}, or MatterPort3D~\cite{Matterport3D}.}
%     \label{fig:comparison_datasets}
% \end{figure}

We set up scenes composed of multiple objects with different shapes and materials to analyse sensor characteristics. 3D models of photometrically challenging objects with reflective or transparent surfaces are recorded with high quality a priori and aligned to the scenes. Images are captured from a synchronised multi-modal custom sensor mounted at a robot end-effector to allow for precise pose camera measurements~\cite{PhoCal}. High-quality rendered depth can be extracted a posteriori from the fully annotated scenes for the viewpoint of each sensor. The acquisition pipeline is depicted in Fig.~\ref{fig:annotation_overview}.
 
Previous 3D and depth acquisition setups~\cite{replica19arxiv,Matterport3D,dai2017scannet} scan the scene as a whole which limits the quality by the used sensor. We instead separately scan every single object, including chairs and background, as well as small household objects a priori with two high-quality structured light object scanners. 
This process significantly pushes the annotation quality for the scenes as the robotic 3D labelling process only has a point RMSE error of $0.80$~mm~\cite{PhoCal}. 
For comparison, a Kinect Azure camera induces a standard deviation of $17$~mm in its working range~\cite{liu2021stereobj}. 
The accuracy allows us to investigate depth errors arising from sensor noise objectively, as shown in Fig.~\ref{fig:dataset_quality}, while resolving common issues of imperfect meshes in available datasets (cf. Fig.~\ref{fig:teaser}, left).

\subsection{Sensor Setup \& Hardware Description}
\begin{figure}[!t]
 \centering
    \includegraphics[width=0.9\linewidth]{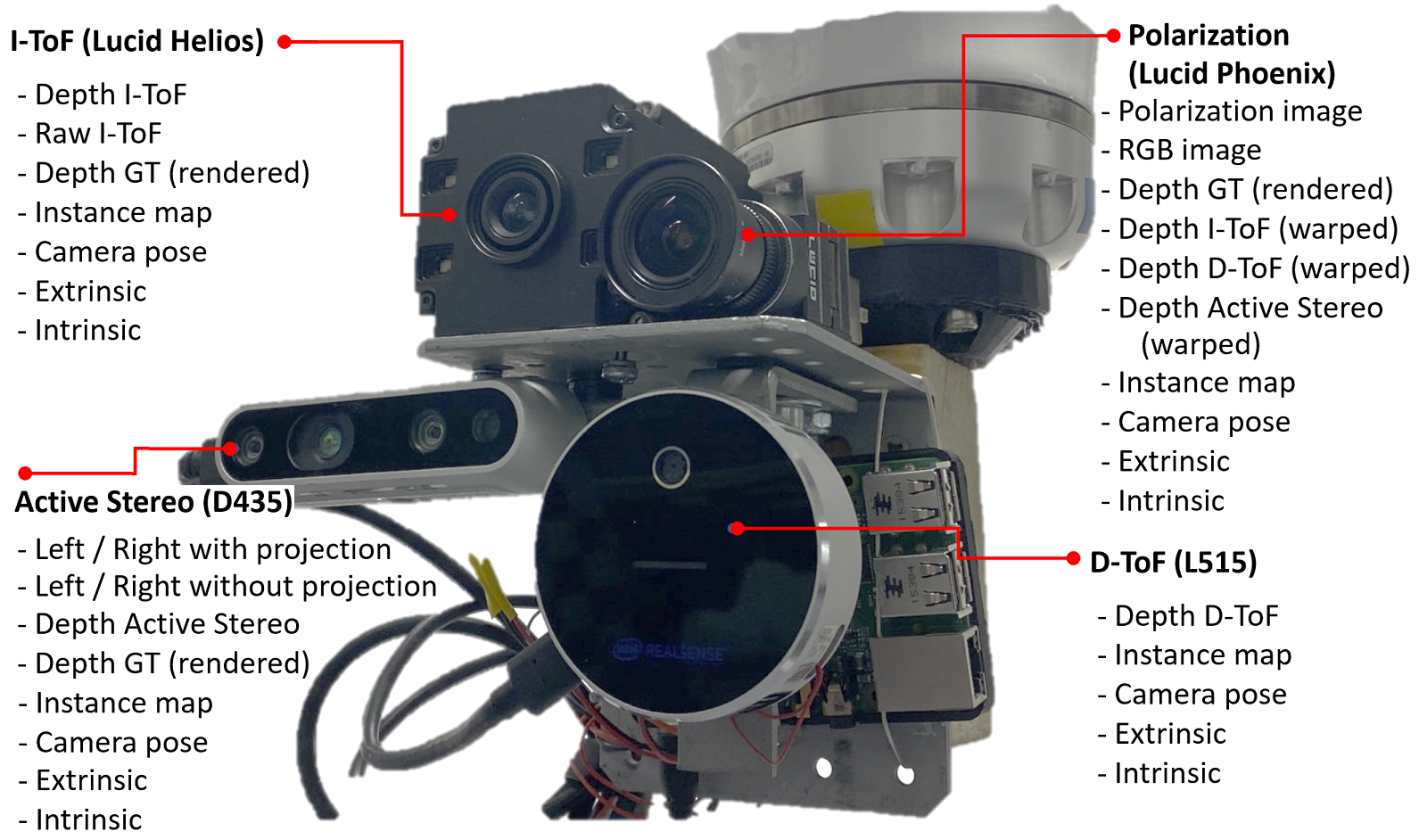}
    \vspace{-2mm}
    \caption{\textbf{Camera Rig and 3D Sensor Data.} The custom multi-modal sensor rig comprises depth sensors for I-ToF (top left), Stereo (lower left), D-ToF (lower right), and RGB-P (Polarization, top right). It is fixed to a robot end-effector (top) and a Raspberry Pi (right) triggers acquisition.}
    \label{fig:hardware}
\end{figure}

The table-top scanner (EinScan-SP, SHINING 3D Tech. Co., Ltd., Hangzhou, China) uses a rotating table and is designed for small objects. The other is a hand-held scanner (Artec Eva, Artec 3D, Luxembourg) which we use for larger objects and the background. 
For objects and areas with challenging material, self-vanishing 3D scanning spray (AESUB Blue) is used. For larger texture-less areas such as tables and walls we temporarily attach small markers~\cite{garrido2014automatic} to the surface to allow for relocalization of the 3D scanner. 
The robotic manipulator is a KUKA LBR iiwa 7 R800 (KUKA Roboter GmbH, Germany) with a position accuracy of $\pm0.1$~mm. 
We validated this during our pivot calibration stage (Fig.~\ref{fig:annotation_overview} b) by calculating the 3D location of the tool tip (using forward kinematics and hand-tip calibration) while varying robot poses. The position varied in $\left[-0.158,0.125\right]$~mm in line with this.
Our dataset features a unique multi-modal setup with four different cameras, which provide four types of input images (RGB, polarization, stereo, Indirect ToF (I-ToF) correlation) and three different depth images modalities (Direct ToF (D-ToF), I-ToF, Active Stereo). RGB and polarization images are acquired with a Phoenix 5.0 MP Polarization camera (PHX050S1-QC, LUCID Vision Labs, Canada) equipped with a Sony Polarsens sensor (IMX264MYR CMOS, Sony, Japan). 
To acquire stereo images, we use an Intel RealSense D435 (Intel, USA) with switched off infrared projector. Depth is acquired from an Intel RealSense L515 D-ToF sensor, an Intel Realsense D435 active stereo sensor with infrared pattern projection, and a Lucid Helios (HLS003S-001, LUCID Vision Labs, Canada) I-ToF sensor. 
A Raspberry Pi triggers each camera separately to remove interference effects between infrared signals of depth sensors. 
The hardware is rigidly mounted at the robot end-effector (see Fig.~\ref{fig:hardware}) which allows to stop frame-by-frame for the synchronized acquisition of a pre-recorded trajectory.

\begin{table*}[!ht]
\setlength{\tabcolsep}{7pt}
\centering
\caption{\textbf{Comparison of Datasets}. Shown are differences between our dataset and previous multi-modal depth datasets for indoor environments. Our dataset is the only one that provides highly accurate GT (Depth, Surface Normals, 6D Object Poses, Instance Masks, Camera Poses, Dense Scene Mesh) together with varying sensor data for real scenes.}
\label{tab:dataset_comparison}
\vspace{-2mm}
\begin{tabular}{r | c c c c c c c c c c c c } 
\toprule
  \small Dataset &
  {\small Acc.GT}          &
  {\small RGB}        &
  {\small D-ToF}          &
  {\small I-ToF}        &
  {\small Stereo} &
  {\small Act.Stereo} &
  {\small Polar.} &
  {\small Indoor}          &
  {\small Real}         &
  {\small Video}        &
  {\small Frames}       \\ 
\midrule
Agresti~\cite{Agresti_2019_CVPR} & -          &  - & - & \checkmark & -          & -& -          & \checkmark & \checkmark & -          & $113$      \\
CroMo~\cite{CroMo} & - &  -  & - & \checkmark & \checkmark & \checkmark & \checkmark & (\checkmark) & \checkmark & \checkmark & ${>}10$k    \\
Zhu~\cite{zhu2019depth}      & -            & (\checkmark) & -          & - & - & - & \checkmark & \checkmark & \checkmark & -        & $1$   \\
Sturm~\cite{sturm2012benchmark}         & - & \checkmark & - & - & - & -          & -          & \checkmark & \checkmark & \checkmark & ${>}10$k   \\
\cite{kadambi2017depth}/\cite{qiu:2019a}/\cite{ba2020deep}            & - & \checkmark   & -          & - & - & - & \checkmark & \checkmark & \checkmark & -          & $1/40/300$ \\
Guo~\cite{Guo_2018_ECCV}       & \checkmark   & -          & - & \checkmark & -          & - & -          & \checkmark & - & - & $2000$\\
\textbf{Ours} & \checkmark & \checkmark & \checkmark & \checkmark & \checkmark & \checkmark & \checkmark & \checkmark & \checkmark & \checkmark & ${>}10$k\\
\bottomrule
\end{tabular}
\end{table*}

\subsection{Scene Statistics \& Data Comparison}
We scanned 7 indoor areas, 6 tables, and 4 chairs, with the handheld scanner as background and large objects. 64 household objects from 9 categories (bottle, can, cup, cutlery, glass, remote, teapot, tube, shoe) are scanned with the tabletop structured light scanner. The data comprises 13 scenes split into 10 scenes for training and 3 scenes for testing. Each scene is recorded with 2 trajectories of 200-300 frames with and without the objects. This sums up to 800-1200 frames per scene, with a total of 10k frames for training and 3k frames for our test set. The 3 test scenes have different background setups: 1) Seen background, 2) Seen background with different lighting conditions and 3) Unseen background and table, with three different object setups: 1) Seen objects 2) Unseen objects from the seen category 3) Unseen objects from unseen categories (shoe and tube).
Table~\ref{tab:dataset_comparison} compares our dataset with various existing setups. To the best of our knowledge, our dataset is the only multi-modal dataset comprising RGB, ToF, Stereo, Active Stereo, and Polarisation modalities simultaneously with reliable ground truth depth maps. 
\section{Methodology}
\label{sec:methods}
The dataset described above allows for the first time for rigorous, in-depth analysis of different depth sensor modalities and a detailed quantitative evaluation of learning-based dense scene regression methods when trained with varying supervision signals.
We focus on the popular tasks of monocular depth estimation and implicit 3D reconstruction with the application of novel view synthesis.

\subsection{Depth Estimation}
To train the depth estimation from a single image, we leverage the widely adopted architecture from~\cite{monodepth2}. 
%This allows to learn with full supervision, perform semi-supervised training with ground-truth camera poses, and train with self-supervised configurations.
We train an encoder-decoder network with a ResNet18 encoder and skip connections to regress dense depth. Using different supervision signals from varying depth modalities allows to study the influence and the characteristics of the 3D sensors. 
Additionally, we analyze whether complementary semi-supervision via information of the relative pose between monocular acquisitions and consecutive image information of the moving camera can overcome sensor issues. 

We further investigate the network design influence on the prediction quality for the supervised case. For this, we train two high-capacity networks with transformer backbones on our data, namely DPT~\cite{ranftl2021vision} and MIDAS~\cite{Ranftl2022}. 

\paragraph{Dense Supervision}
In the fully supervised setup, depth modalities from the dataset are used to supervise the prediction of the four pyramid level outputs after upsampling to the original input resolution with:
%\begin{align}
$\mathcal{L}_{\text{supervised}}=\sum_{i=1}^{i=4}\left\Vert{\widetilde{D}}_i-{D}\right\Vert_{1}$,
  \label{equ:soft-argmin}
%\end{align}
where $D$ is the supervision signal for valid pixels of the depth map and $\widetilde{D}_i$ the predicted depth at pyramid scale $i$.

\paragraph{Self-Supervision}
Depth and relative pose prediction between consecutive frames of a moving camera can be formulated as coupled optimization problem. We follow established methods to formulate a dense image reconstruction loss through projective geometric warping~\cite{monodepth2}.
In this process, a temporal image $I_{t^\prime}$ at time $t^\prime$ is projectively transformed to the frame at time $t$ via:\\
% \begin{align}
$I_{t^\prime \to t} = I_{t^\prime}\Big\langle \text{proj}(D_t, T_{t \to t^\prime}, K) \Big\rangle$,
\label{eqn:warp}
% \end{align}
where $D_t$ is the predicted depth for frame $t$, $T_{t \to t^\prime}$ the relative camera pose, and $K$ the camera intrinsics.
The photometric reconstruction error~\cite{monodepth2,watson2021temporal,ruhkamp2021attention} between image $I_x$ and $I_y$, given by:
%~\cite{Godard_2017}
%\begin{align}
${E_{\text{pe}}}(I_x,I_y) = \alpha\tfrac{1-\text{SSIM}(I_x, I_y)}{2} + (1-\alpha) \left \Vert I_x-I_y \right \Vert_{1}$
\label{eqn:photo}
%\end{align}
is computed between target frame $I_t$ and each source frame $I_s$ with $s \in S$. The pixel-wise minimum error is retrieved to finally define $\mathcal{L}_{\text{photo}}$ over $S = [t-F, t+F]$ as 
% \begin{align}
$\mathcal{L}_{\text{photo}} = \min_{s \in S}  {E_{\text{pe}}}(I_t,I_{s \rightarrow t})$.
% \end{align}
The edge-aware smoothness $\mathcal{L}_{\text{s}}$ is applied~\cite{monodepth2} to encourage locally smooth depth estimations with the mean-normalized inverse depth $\overline{d_{t}}$ as
% \begin{align}
$\mathcal{L}_{\text{s}} = \left|\partial_{x}\overline{d_{t}}\right|e^{-\left|\partial_{x}I_{t}\right|} + \left|\partial_{y}\overline{d_{t}}\right|e^{-\left|\partial_{y}I_{t}\right|}$.
% \end{align}
The final training loss for the self-supervised setup is:
%\begin{align}
$\mathcal{L}_{\text{self-supervised}} = \mathcal{L}_{\text{photo}} +  \lambda_{\text{s}} \cdot \mathcal{L}_{\text{s}}$.
%\end{align}

\paragraph{Semi-Supervision}
For the semi-supervised training, the ground truth relative camera pose is leveraged. 
%to perform differentiable backwards warping with bilinear interpolation. 
The predicted depth estimate is used to formulate the photometric image reconstruction. We also enforce the smoothness loss as detailed above.

\paragraph{Data Fusion}
Despite providing high accuracy ground truth, our annotation pipeline is time-consuming. 
One may ask whether this cannot be done with multi-view data aggregation. 
We therefore compare the quality against the dense structure from motion method Kinect Fusion~\cite{kinectfusion} and an approach for TSDF Fusion~\cite{Zhou2018}. 
The synchronized sensor availability allows also to investigate and improve sensor fusion pipelines. To illustrate the impact of high quality GT for this task, we also train the recent raw ToF+RGB fusion network Wild-ToFu~\cite{jung2021wild} on our dataset. 

\subsection{Implicit 3D Reconstruction}
%\pat{describe problem, NeRF stuff + some details on DS-NeRF regularization}
Recent work on implicit 3D scene reconstruction leverages neural radiance fields (NeRF)~\cite{mildenhall2021nerf}. 
The technique works particularly well for novel view synthesis and allows to render scene geometry or RGB views from unobserved viewpoints.
%with a set of input images with leverages known camera intrinsics and extrinsics through differentiable rendering techniques. 
%While the focus of applications is to render RGB images from the implicit scene representation, a large capacity of the network weights encode the 3D scene geometry, as this is vital for accurate novel view synthesis. 
Providing additional depth supervision regularizes the problem such that fewer views are required and training efficiency is increased~\cite{deng2022depth,roessle2022dense}.
%By providing an additional supervision signal for the synthesized depth representation, e.g. from sparse structure-from-motion (SfM) estimates, the optimisation can be improved and less views are required~\cite{deng2022depth}.
%When providing dense depth priors for the neural radiance field~\cite{roessle2022dense}, novel view synthesis and geometric scene estimates can be improved even more. 
%In~\cite{roessle2022dense} dense depth priors are learned through depth completion of sparse SfM 3D points by supervising with RGB-D data for later use in the NeRF.
%This does not take into account the inherent artefacts of such sensors as discussed here.
We follow the motivation of~\cite{roessle2022dense} and leverage different depth modalities to serve as additional depth supervision for novel view synthesis. 
Following NeRF literature~\cite{mildenhall2021nerf,roessle2022dense}, we encode the radiance field for a scene in an MLP $F_{\theta}$ to predict colour $\mathbf{C} = [r, g, b]$ and volume density $\sigma$ for some 3D position $\mathbf{x} \in \mathbb{R}^3$ and viewing direction $\mathbf{d} \in \mathbb{S}^2$. We use the positional encoding from~\cite{roessle2022dense}.
For each pixel, a ray $\mathbf{r}(t) = \mathbf{o} + t\mathbf{d}$ from the camera origin $\mathbf{o}$ is sampled through the volume at location $t_k \in [t_n, t_f]$ between near and far planes by querying $F_{\theta}$ to obtain colour and density:\\
% \begin{align}
$\hat{\mathbf{C}}(\mathbf{r}) = \sum_{k=1}^{K} w_k\mathbf{c}_k \hspace{0.25em} \text{with} \hspace{0.25em}  w_k=T_k\left(1 - \exp(-\sigma_k\delta_k)\right)$, \\
%\, , 
% \end{align}
% \begin{align}
$T_k= \exp \left( -\sum_{k'=1}^{k} \sigma_{k'}\delta_{k'}\right)\, \hspace{0.25em} \text{and}  \hspace{0.25em}
\delta_k = t_{k+1} - t_k$. \\
%\, .
% \end{align}
The NeRF depth $\hat{z}(\mathbf{r})$ is computed by:
%\begin{align}
$\hat{z}(\mathbf{r}) = \sum_{k=1}^{K}w_kt_k$
%\end{align}
% NeRF parameters $\theta$ are optimized via the standard mean squared error (MSE)loss for color $\mathcal{L}_{\mathrm{color}}$ 
and the depth regularization for an image with rays $\mathcal{R}$ is:
% $\mathcal{L}_{\mathrm{depth}}$: 
%\begin{align}
%\mathcal{L}_{\text{col}} = \displaystyle\sum_{\mathbf{r}}
%\begin{Vmatrix} \hat{\mathbf{C}}(\mathbf{r}) - \mathbf{C}(\mathbf{r}) \end{Vmatrix}_2^2, \ 
 %    + 
%     \lambda \frac{\vert\hat{z}(\mathbf{r}) - {z}(\mathbf{r})\vert}{\hat{z}(\mathbf{r}) + {z}(\mathbf{r})}\Big) , 
    % \mathcal{L}_{\mathrm{color}}(\mathbf{r}) + \lambda \mathcal{L}_{\mathrm{depth}}(\mathbf{r})
    % \\
    % \mathcal{L}_{\mathrm{color}}(\mathbf{r}) &= \begin{Vmatrix} \hat{\mathbf{C}}(\mathbf{r}) - \mathbf{C}(\mathbf{r}) \end{Vmatrix}_2^2, \\
    % \mathcal{L}_{\mathrm{depth}}(\mathbf{r}) &= \frac{\vert\hat{z}(\mathbf{r}) - {z}(\mathbf{r})\vert}{\hat{z}(\mathbf{r}) + {z}(\mathbf{r})}, 
$\mathcal{L}_{\text{D}} = \displaystyle\sum_{\mathbf{r} \in \mathcal{R}}
%\begin{Vmatrix} \hat{\mathbf{C}}(\mathbf{r}) - \mathbf{C}(\mathbf{r}) \end{Vmatrix}_2^2
 %    + 
     \frac{\vert\hat{z}(\mathbf{r}) - {z}(\mathbf{r})\vert}{\hat{z}(\mathbf{r}) + {z}(\mathbf{r})}$ , 
%\end{align}
where $z(\mathbf{r})$ is the depth of the sensor.
%For more details on ray sampling, depth-guided sampling and optimisation, we refer to~\cite{roessle2022dense}.
Using the mean squared error (MSE) loss 
$\mathcal{L}_{\text{colour}} = \text{MSE}(\hat{\mathbf{C}}, {\mathbf{C}})$ 
% $\mathcal{L}_{\text{colour}} = \begin{Vmatrix}\hat{\mathbf{C}}, {\mathbf{C}}\end{Vmatrix}_2^2$ 
for synthesized colours, the final training loss is:
%\begin{align}
$\mathcal{L}_{\text{NeRF}} = \mathcal{L}_{\text{colour}} +  \lambda_{\text{D}} \cdot \mathcal{L}_{\text{D}}$.
%\end{align}
\section{Sensor Impact for Dense 3D Vision Tasks}
\label{sec:experiments}

\begin{table}[!b]
\centering
\caption{\textbf{Depth Prediction Results for Different Training Signals.} Top: Dense supervision from different depth modalities. Bottom: Evaluation of semi-supervised (pose GT) and self-supervised (mono and mono+stereo) training. The entire scene (Full), background (BG), and objects (Obj) are evaluated separately. Objects material is further split into textured, reflective and transparent. \textbf{Best} and \underline{2nd best} RMSE in mm are indicated.
}
\footnotesize
\resizebox{\columnwidth}{!}{
\begin{tabular}{ll|c|cc|ccc}
\toprule
& Training Signal &\multicolumn{1}{c|}{Full} & BG & \multicolumn{1}{c|}{Obj}  & Text. & Refl. & Transp. \\
\midrule
\multirow{3}{*}{\rotatebox[origin=c]{90}{Sup.}}     & I-ToF & 113.29 & 111.13 & 119.72 & 54.45 & 87.84 & 207.89 \\
                                & D-ToF & \underline{77.97} & \textbf{69.87}& \underline{112.83}& \textbf{37.88} & \underline{71.59} & \underline{207.85} \\
                                & Active Stereo & \textbf{72.20} & \underline{71.94}& \textbf{61.13}& \underline{50.90} & \textbf{52.43} & \textbf{87.24} \\
\midrule\\
\midrule
\multirow{3}{*}{\rotatebox[origin=c]{90}{Sel/Sem}}     & Pose & \textbf{154.87} & \textbf{158.67} & \textbf{65.42}& \textbf{57.22} & \textbf{37.78} & \underline{61.86} \\
& M & 180.34 & 183.65& 85.51 & 84.26 & \underline{48.80} & \textbf{49.62} \\
                                & M+S & \underline{159.80} & \underline{161.65} & \underline{82.16} &\underline{71.24} & 63.92 & 66.48\\                        
\bottomrule
\end{tabular}
}
\label{tab:depth_supervision_results}
\end{table}

We train a series of networks for the task of monocular depth estimation and implicit scene reconstruction.

\subsection{Depth Estimation}
\begin{figure}[!t]
 \centering
    \includegraphics[width=\linewidth]{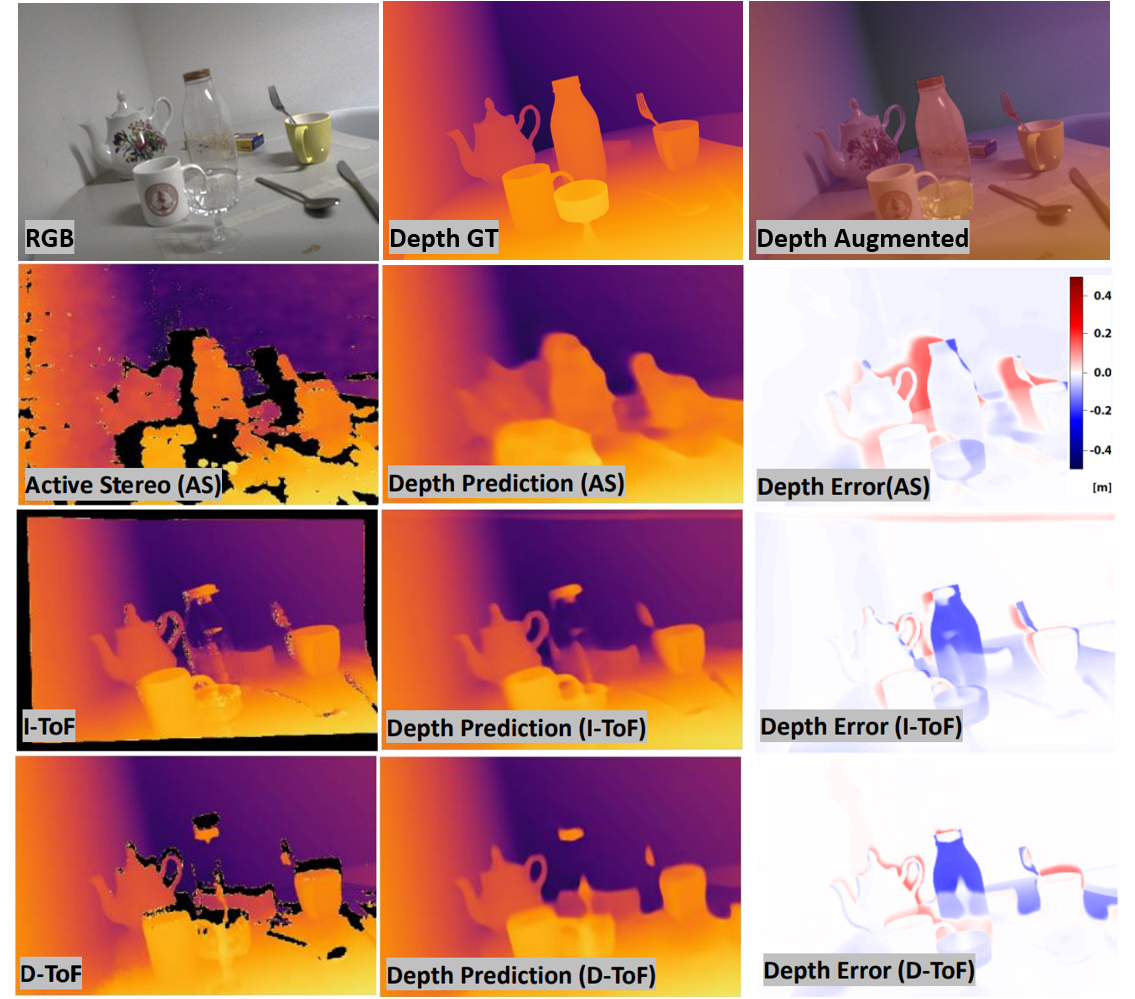}
    \vspace{-6mm}
    \caption{\textbf{Fully Supervised Monocular Depth.} Monocular depth tends to overfit on the specific noise of the sensor the network is trained on. Prediction from Active Stereo GT is robust on the material while depth map is blurry, while both I-ToF and D-ToF has strong material dependent artifact but sharp on the edges.}
    \label{fig:depth_qualitative}
\end{figure}

Results for monocular depth estimation with varying training signal are summarized in Table~\ref{tab:depth_supervision_results} and Fig.~\ref{fig:depth_qualitative}. 
We report average results for the scenes and separate performances for background, objects, and materials of different photometric complexity. 
The error varies from background to objects. Their varying photometric complexity can explain this. Not surprisingly, the ToF training is heavily influenced by reflective and transparent object material, where the active stereo camera can project some patterns onto diffusely reflective surfaces. 
Interestingly, the self- and semi-supervised setups help to recover information in these challenging setups to some extent, such that these cases even outperform the ToF supervision for photometrically challenging objects. In contrast, simpler structures (such as the background) benefit from the ToF supervision. 
This indicates that sensor-specific noise is learnt and reveals that systematic errors of learning approaches cannot be evaluated if such 3D devices are used for ground truth evaluation without critical analysis. 
This might ultimately lead to incorrect result interpretations, particularly if self-supervised approaches are evaluated against co-modality sensor data.
The table also discloses that the mutual prediction of inter-frame poses in self-supervision indoor setups is challenging, and accurate pose labels can have an immediate and significant impact on the depth results (Pose vs. M).

\begin{figure}[!t]
 \centering
    \includegraphics[width=\linewidth]{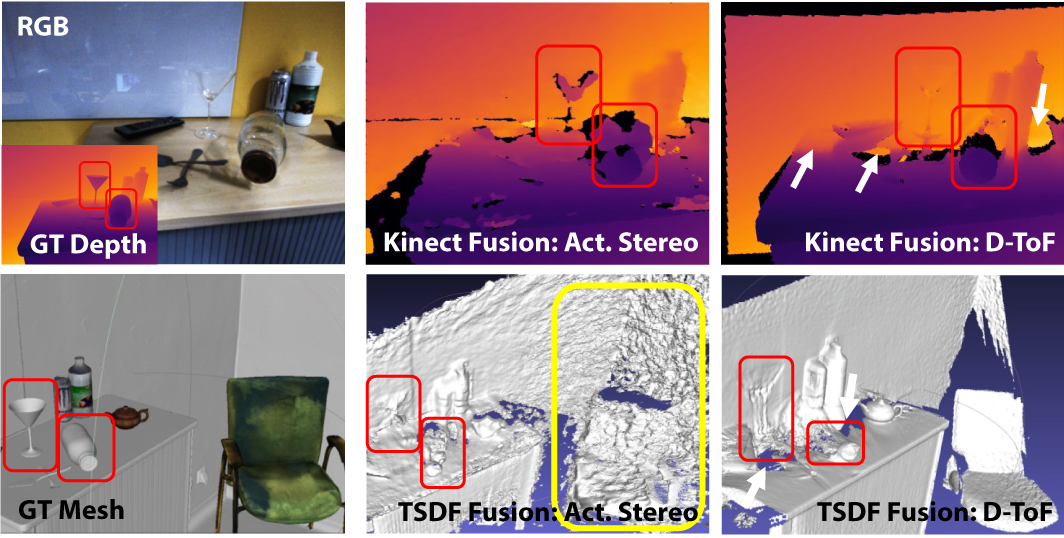}
    \vspace{-6mm}
    \caption{\textbf{Dense SfM.} A scene with our GT (left), Kinect~\cite{kinectfusion} (top) and TSDF~\cite{Zhou2018} (bottom) fusion approaches. Inherent sensor noise due to MPI (white), transparent objects (red), and diffuse texture-less material (yellow) persists.}
    \label{fig:fusion}
\end{figure}

\begin{figure}[!b]
 \centering
    \includegraphics[width=\linewidth]{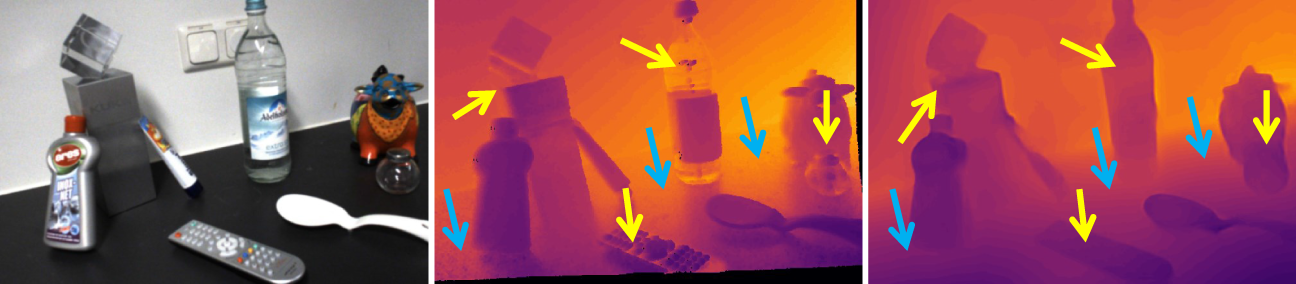}
    \vspace{-6mm}
    \caption{\textbf{Sensor Fusion.} Scene (left) with I-ToF depth (centre) and ToF+RGB Fusion~\cite{jung2021wild} (right). Fusion can help to resolve some material induced artefacts (yellow) as well as MPI (blue).}
    \label{fig:wildtofu}
\end{figure}

\begin{figure*}[!t]
 \centering
    \includegraphics[width=1.0\linewidth]{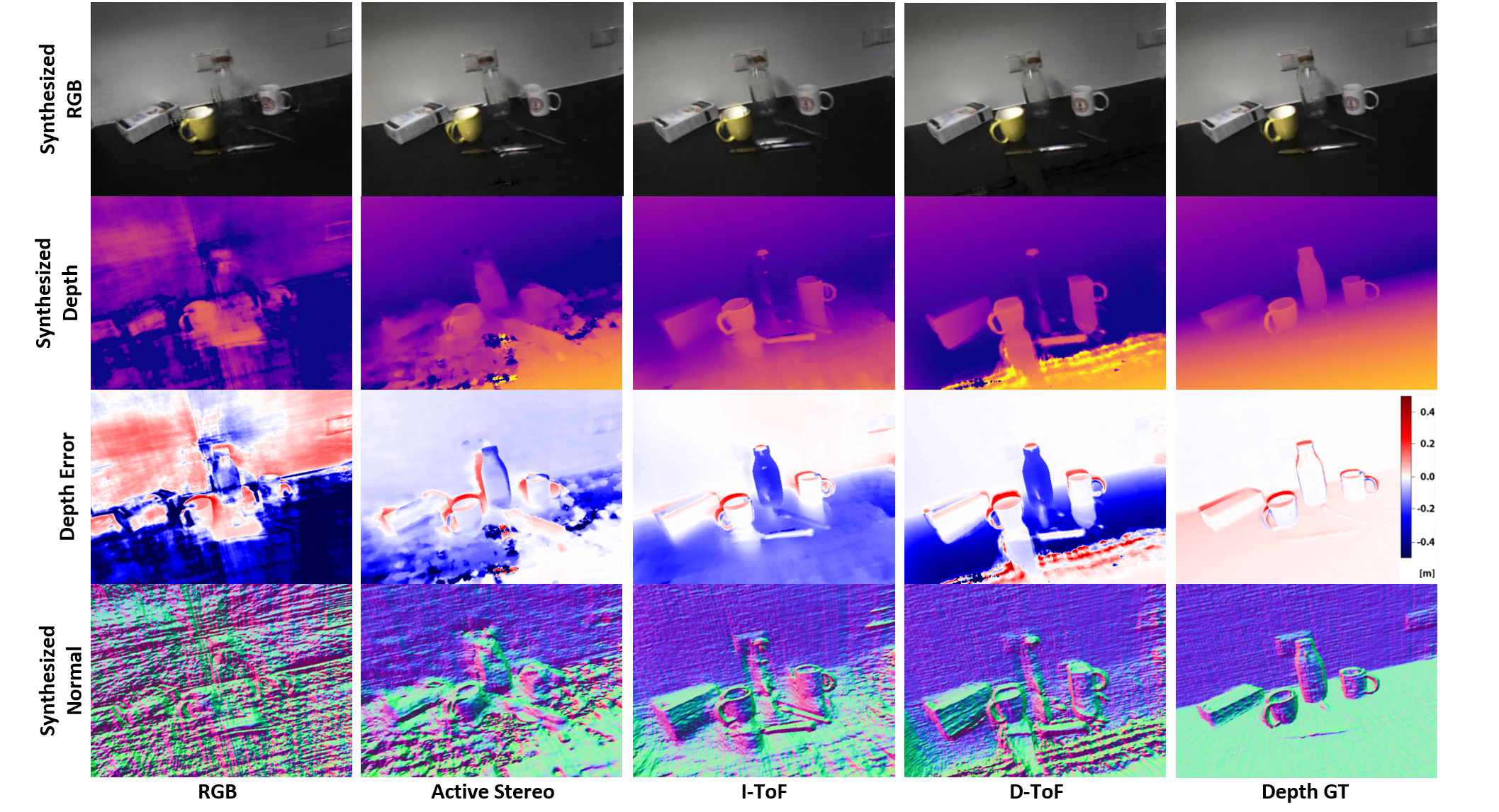}
    \vspace{-7mm}
    \caption{\textbf{Reconstruction Results.} The results of an implicit scene reconstruction with a Neural Radiance Field (NeRF) are shown. Images are synthesised for depth, surface normals and RGB for an unseen view, which is shown together with the prediction errors. The columns allow us to compare different methods where a NeRF~\cite{mildenhall2021nerf} is trained solely on RGB (first column) and various depth maps for regularisation as proposed in~\cite{roessle2022dense}. The last column illustrates synthesised results from training with GT depth for comparison. Differences are visible, especially for the partly reflective table edges, the translucent bottle and around depth discontinuities.}
    \label{fig:nerf_comparison}
\end{figure*}

Fig.~\ref{fig:fusion} shows that multi-view data aggregation in the form of dense SfM fails to reproduce highly reliable 3D reconstructions. In particular transparent and diffuse texture-less objects pose challenges to both Active Stereo and D-ToF. These can neither be recovered by the Kinect Fusion pipeline~\cite{kinectfusion} nor by the TSDF Fusion implementation of Open3D~\cite{Zhou2018} for which we use the GT camera poses. Inherent sensor artefacts are present even if depth maps from different viewpoints are combined. This quality advantage justifies our expensive annotation setup.
We further analysed the results of training runs with DPT~\cite{ranftl2021vision} and MIDAS~\cite{Ranftl2022}, which we train from scratch. While these more complex architectures with higher capacity show the same trend and also learn sensor noise, the training time is significantly longer. More details are provided in the supplementary material.
From the previous results, we have seen that ToF depth is problematic for translucent and reflective material. Fig~\ref{fig:wildtofu} illustrates that an additional co-modal input signal at test time can cure these effects partly. It can be observed that the use of additional RGB data in~\cite{jung2021wild} reduces the influence of MPI and resolves some material-induced depth artefacts. Our unique dataset also inspires cross-modal fusion pipelines' development and objective analysis.

\subsection{Implicit 3D Reconstruction {\&} View Synthesis}
Our implicit 3D reconstruction generates novel views for depth, normals and RGB with varying quality. If trained with only colour information, the NeRF produces convincing RGB views with the highest PSNR (cf. Fig.~\ref{fig:nerf_comparison} and Table~\ref{tab:nerf_results}).
However, the 3D scene geometry is not well reconstructed.
In line with the literature~\cite{deng2022depth,roessle2022dense}, depth regularization improves this (e.g. on texture-less regions).
Regularising with different depth modalities makes the sensor noise of I-ToF, AS, and D-ToF clearly visible. 
While the RMSE behaves similarly to the monocular depth prediction results with AS as best, followed by D-ToF and I-ToF. The cosine similarity for surface normal estimates confirms this trend.
The overall depth and normal reconstruction for AS are very noisy, but depth error metrics are more sensitive for significant erroneous estimates for reflective and translucent objects. 
Prior artefacts of the respective sensor influence the NeRF and translate into incorrect reconstructions (e.g. errors from D-ToF and I-ToF for translucent material or noisy background and inaccurate depth discontinuities at edges for AS).
Interestingly, the D-ToF prior can improve the overall reconstruction for most of the scene but fails for the bottle, where the AS can give better depth priors. This is also visible in the synthesised depth.
Leveraging synthetic depth GT (last row) mitigates these issues and positively affects the view synthesis with higher SSIM.

\begin{table}[!b]
\centering
\caption{\textbf{Novel View Synthesis from Implicit 3D Reconstruction.} Evaluation against GT for RGB, depth and surface normal estimates for different optimisation strategies (RGB-only for supervision and $+$ respective sensor depth). We indicate \textbf{best}, \underline{2nd best} and \dashuline{3rd best}. Depth metrics in mm.
}
\vspace{-1mm}
\footnotesize
\resizebox{\columnwidth}{!}{
\begin{tabular}{l|cc|cccc|c}
\toprule
& \multicolumn{2}{c|}{RGB}
& \multicolumn{4}{c|}{Depth}
& Normal
\\
Modality & PSNR $\uparrow$ & SSIM $\uparrow$ & Abs.Rel.$\downarrow$ &   Sq.Rel.$\downarrow$ &     RMSE $\downarrow$ & 
$\sigma < 1.25$ $\uparrow$
& Cos.Sim.$\downarrow$

\\

\midrule
RGB Only&
\textbf{32.406}&
\underline{0.889}
&   0.328  &   111.229  & 226.187  &
0.631  
& 0.084
\\
\midrule
$+$ AS&
17.570&
0.656
&   \dashuline{0.113}  &   \underline{16.050}  &  \underline{94.520}  &    
\dashuline{0.853}     
& \dashuline{0.071}
\\

$+$ I-ToF&
18.042&
0.653
&   0.296  &   91.426  & 217.334  &
0.520     
& 0.102
\\
$+$ D-ToF&
\dashuline{31.812} &
\dashuline{0.888}
&   \underline{0.112}  &  \dashuline{24.988}  & \dashuline{119.455}  &
\underline{0.882}
& \underline{0.031}
\\
$+$ Syn. &
\underline{32.082}&
\textbf{0.894}
&   \textbf{0.001}  &   \textbf{0.049}  &   \textbf{3.520}  &   
\textbf{1.000}     
& \textbf{0.001}
\\

\bottomrule
\end{tabular}
}
\label{tab:nerf_results}
\end{table}
\section{Discussion \& Conclusion}
\label{sec:discussion}
This paper shows that questioning and investigating commonly used 3D sensors helps to understand their impact on dense 3D vision tasks.
For the first time, we make it possible to study how sensor characteristics influence learning in these areas objectively. We quantify the effect of various photometric challenges, such as translucency and reflectivity for depth estimation, reconstruction and novel view synthesis and provide a unique dataset to stimulate research in this direction.
While obvious sensor noise is not "surprising", our dataset quantifies this impact for the first time.
For instance, interestingly, D-ToF supervision is significantly better suited (13.02~mm) for textured objects than AS, which in return surpasses I-ToF by 3.55~mm RMSE (cf.~\ref{tab:depth_supervision_results}). Same trend holds true on mostly texture-less backgrounds where D-ToF is 37\% more accurate than I-ToF.
For targeted analysis and research of dense methods for reflective and transparent objects, a quantitative evaluation is of utmost interest
- while our quantifiable error maps allow specifying the detailed deviations.
Although our dataset tries to provide scenes with varying backgrounds, the possible location of the scene is restricted due to the limited working range of the robot manipulator.
Aside from our investigations and the evaluation of sensor signals for standard 3D vision tasks, we firmly believe that our dataset can also pave the way for further investigation of cross-modal fusion pipelines.

\clearpage
\newpage
%%%%%%%%% REFERENCES
{\small
\bibliographystyle{ieee_fullname}
\bibliography{PaperForReview}

\begin{thebibliography}{10}\itemsep=-1pt

\bibitem{aanaes2016IJCV}
Henrik Aanæs, Rasmus Jensen, George Vogiatzis, Engin Tola, and Anders Dahl.
\newblock Large-scale data for multiple-view stereopsis.
\newblock {\em International Journal of Computer Vision}, 120, 11 2016.

\bibitem{Agresti_2019_CVPR}
Gianluca Agresti, Henrik Schaefer, Piergiorgio Sartor, and Pietro Zanuttigh.
\newblock Unsupervised domain adaptation for tof data denoising with
  adversarial learning.
\newblock In {\em Proceedings of the IEEE/CVF Conference on Computer Vision and
  Pattern Recognition (CVPR)}, June 2019.

\bibitem{atkinson2006recovery}
Gary~A Atkinson and Edwin~R Hancock.
\newblock Recovery of surface orientation from diffuse polarization.
\newblock {\em IEEE transactions on image processing}, 15(6):1653--1664, 2006.

\bibitem{ba2020deep}
Yunhao Ba, Alex Gilbert, Franklin Wang, Jinfa Yang, Rui Chen, Yiqin Wang, Lei
  Yan, Boxin Shi, and Achuta Kadambi.
\newblock Deep shape from polarization.
\newblock In {\em Computer Vision--ECCV 2020: 16th European Conference,
  Glasgow, UK, August 23--28, 2020, Proceedings, Part XXIV 16}, pages 554--571.
  Springer, 2020.

\bibitem{barron2022mip}
Jonathan~T Barron, Ben Mildenhall, Dor Verbin, Pratul~P Srinivasan, and Peter
  Hedman.
\newblock Mip-nerf 360: Unbounded anti-aliased neural radiance fields.
\newblock In {\em Proceedings of the IEEE/CVF Conference on Computer Vision and
  Pattern Recognition}, pages 5470--5479, 2022.

\bibitem{busam2019sterefo}
Benjamin Busam, Matthieu Hog, Steven McDonagh, and Gregory Slabaugh.
\newblock {SteReFo}: efficient image refocusing with stereo vision.
\newblock In {\em Proceedings of the IEEE/CVF International Conference on
  Computer Vision Workshops}, 2019.

\bibitem{Butler:ECCV:2012}
D.~J. Butler, J. Wulff, G.~B. Stanley, and M.~J. Black.
\newblock A naturalistic open source movie for optical flow evaluation.
\newblock In {A. Fitzgibbon et al. (Eds.)}, editor, {\em European Conf. on
  Computer Vision (ECCV)}, Part IV, LNCS 7577, pages 611--625. Springer-Verlag,
  Oct. 2012.

\bibitem{Matterport3D}
Angel Chang, Angela Dai, Thomas Funkhouser, Maciej Halber, Matthias Niessner,
  Manolis Savva, Shuran Song, Andy Zeng, and Yinda Zhang.
\newblock Matterport3d: Learning from rgb-d data in indoor environments.
\newblock {\em International Conference on 3D Vision (3DV)}, 2017.

\bibitem{chen2022tensorf}
Anpei Chen, Zexiang Xu, Andreas Geiger, Jingyi Yu, and Hao Su.
\newblock Tensorf: Tensorial radiance fields.
\newblock {\em arXiv preprint arXiv:2203.09517}, 2022.

\bibitem{chen2019towards}
Po-Yi Chen, Alexander~H Liu, Yen-Cheng Liu, and Yu-Chiang~Frank Wang.
\newblock Towards scene understanding: Unsupervised monocular depth estimation
  with semantic-aware representation.
\newblock In {\em Proceedings of the IEEE Conference on Computer Vision and
  Pattern Recognition}, pages 2624--2632, 2019.

\bibitem{curless1996volumetric}
Brian Curless and Marc Levoy.
\newblock A volumetric method for building complex models from range images.
\newblock In {\em Proceedings of the 23rd annual conference on Computer
  graphics and interactive techniques}, pages 303--312, 1996.

\bibitem{dai2017scannet}
Angela Dai, Angel~X. Chang, Manolis Savva, Maciej Halber, Thomas Funkhouser,
  and Matthias Nie{\ss}ner.
\newblock Scannet: Richly-annotated 3d reconstructions of indoor scenes.
\newblock In {\em Proc. Computer Vision and Pattern Recognition (CVPR), IEEE},
  2017.

\bibitem{deng2022depth}
Kangle Deng, Andrew Liu, Jun-Yan Zhu, and Deva Ramanan.
\newblock Depth-supervised nerf: Fewer views and faster training for free.
\newblock In {\em Proceedings of the IEEE/CVF Conference on Computer Vision and
  Pattern Recognition}, pages 12882--12891, 2022.

\bibitem{eigen2014depth}
David Eigen, Christian Puhrsch, and Rob Fergus.
\newblock Depth map prediction from a single image using a multi-scale deep
  network.
\newblock In {\em Advances in neural information processing systems}, pages
  2366--2374, 2014.

\bibitem{fang2020graspnet}
Hao-Shu Fang, Chenxi Wang, Minghao Gou, and Cewu Lu.
\newblock Graspnet-1billion: A large-scale benchmark for general object
  grasping.
\newblock In {\em Proceedings of the IEEE/CVF Conference on Computer Vision and
  Pattern Recognition}, pages 11444--11453, 2020.

\bibitem{fridovich2022plenoxels}
Sara Fridovich-Keil, Alex Yu, Matthew Tancik, Qinhong Chen, Benjamin Recht, and
  Angjoo Kanazawa.
\newblock Plenoxels: Radiance fields without neural networks.
\newblock In {\em Proceedings of the IEEE/CVF Conference on Computer Vision and
  Pattern Recognition}, pages 5501--5510, 2022.

\bibitem{gao2021polarimetric}
Daoyi Gao, Yitong Li, Patrick Ruhkamp, Iuliia Skobleva, Magdalena Wysocki,
  HyunJun Jung, Pengyuan Wang, Arturo Guridi, and Benjamin Busam.
\newblock Polarimetric pose prediction.
\newblock In {\em European Conference on Computer Vision}, pages 735--752.
  Springer, 2022.

\bibitem{garcia2015surface}
N~Missael Garcia, Ignacio De~Erausquin, Christopher Edmiston, and Viktor Gruev.
\newblock Surface normal reconstruction using circularly polarized light.
\newblock {\em Optics express}, 23(11):14391--14406, 2015.

\bibitem{garg2016unsupervised}
Ravi Garg, Vijay~Kumar Bg, Gustavo Carneiro, and Ian Reid.
\newblock Unsupervised cnn for single view depth estimation: Geometry to the
  rescue.
\newblock In {\em European conference on computer vision}, pages 740--756.
  Springer, 2016.

\bibitem{garrido2014automatic}
Sergio Garrido-Jurado, Rafael Mu{\~n}oz-Salinas, Francisco~Jos{\'e}
  Madrid-Cuevas, and Manuel~Jes{\'u}s Mar{\'\i}n-Jim{\'e}nez.
\newblock Automatic generation and detection of highly reliable fiducial
  markers under occlusion.
\newblock {\em Pattern Recognition}, 47(6):2280--2292, 2014.

\bibitem{gasperini2021r4dyn}
Stefano Gasperini, Patrick Koch, Vinzenz Dallabetta, Nassir Navab, Benjamin
  Busam, and Federico Tombari.
\newblock R4dyn: Exploring radar for self-supervised monocular depth estimation
  of dynamic scenes.
\newblock In {\em 2021 International Conference on 3D Vision (3DV)}, pages
  751--760. IEEE, 2021.

\bibitem{Geiger_2013}
A Geiger, P Lenz, C Stiller, and R Urtasun.
\newblock Vision meets robotics: The kitti dataset.
\newblock {\em The International Journal of Robotics Research},
  32(11):1231–1237, Aug 2013.

\bibitem{geiger2012we}
Andreas Geiger, Philip Lenz, and Raquel Urtasun.
\newblock Are we ready for autonomous driving? the kitti vision benchmark
  suite.
\newblock In {\em 2012 IEEE conference on computer vision and pattern
  recognition}, pages 3354--3361. IEEE, 2012.

\bibitem{Godard_2017}
Clement Godard, Oisin~Mac Aodha, and Gabriel~J. Brostow.
\newblock Unsupervised monocular depth estimation with left-right consistency.
\newblock {\em 2017 IEEE Conference on Computer Vision and Pattern Recognition
  (CVPR)}, Jul 2017.

\bibitem{monodepth2}
Cl{\'{e}}ment Godard, Oisin {Mac Aodha}, Michael Firman, and Gabriel~J.
  Brostow.
\newblock Digging into self-supervised monocular depth prediction.
\newblock In {\em The International Conference on Computer Vision (ICCV)},
  2019.

\bibitem{Guo_2018_ECCV}
Qi Guo, Iuri Frosio, Orazio Gallo, Todd Zickler, and Jan Kautz.
\newblock Tackling 3d tof artifacts through learning and the flat dataset.
\newblock In {\em The European Conference on Computer Vision (ECCV)}, September
  2018.

\bibitem{jung2021wild}
HyunJun Jung, Nikolas Brasch, Ale{\v{s}} Leonardis, Nassir Navab, and Benjamin
  Busam.
\newblock Wild tofu: Improving range and quality of indirect time-of-flight
  depth with rgb fusion in challenging environments.
\newblock In {\em 2021 International Conference on 3D Vision (3DV)}, pages
  239--248. IEEE, 2021.

\bibitem{kadambi2017depth}
Achuta Kadambi, Vage Taamazyan, Boxin Shi, and Ramesh Raskar.
\newblock Depth sensing using geometrically constrained polarization normals.
\newblock {\em International Journal of Computer Vision}, 125(1-3):34--51,
  2017.

\bibitem{kalra2020deep}
Agastya Kalra, Vage Taamazyan, Supreeth~Krishna Rao, Kartik Venkataraman,
  Ramesh Raskar, and Achuta Kadambi.
\newblock Deep polarization cues for transparent object segmentation.
\newblock In {\em Proceedings of the IEEE/CVF Conference on Computer Vision and
  Pattern Recognition}, pages 8602--8611, 2020.

\bibitem{kong2020semantic}
Xin Kong, Xuemeng Yang, Guangyao Zhai, Xiangrui Zhao, Xianfang Zeng, Mengmeng
  Wang, Yong Liu, Wanlong Li, and Feng Wen.
\newblock Semantic graph based place recognition for 3d point clouds.
\newblock In {\em 2020 IEEE/RSJ International Conference on Intelligent Robots
  and Systems (IROS)}, pages 8216--8223. IEEE, 2020.

\bibitem{kopf2021robust}
Johannes Kopf, Xuejian Rong, and Jia-Bin Huang.
\newblock Robust consistent video depth estimation.
\newblock In {\em Proceedings of the IEEE/CVF Conference on Computer Vision and
  Pattern Recognition}, pages 1611--1621, 2021.

\bibitem{laina2016deeper}
Iro Laina, Christian Rupprecht, Vasileios Belagiannis, Federico Tombari, and
  Nassir Navab.
\newblock Deeper depth prediction with fully convolutional residual networks.
\newblock In {\em 2016 Fourth international conference on 3D vision (3DV)},
  pages 239--248. IEEE, 2016.

\bibitem{lee2021patch}
Sihaeng Lee, Janghyeon Lee, Byungju Kim, Eojindl Yi, and Junmo Kim.
\newblock Patch-wise attention network for monocular depth estimation.
\newblock In {\em Proceedings of the AAAI Conference on Artificial
  Intelligence}, volume~35, pages 1873--1881, 2021.

\bibitem{lin2021barf}
Chen-Hsuan Lin, Wei-Chiu Ma, Antonio Torralba, and Simon Lucey.
\newblock Barf: Bundle-adjusting neural radiance fields.
\newblock In {\em Proceedings of the IEEE/CVF International Conference on
  Computer Vision}, pages 5741--5751, 2021.

\bibitem{liu2021stereobj}
Xingyu Liu, Shun Iwase, and Kris~M Kitani.
\newblock Stereobj-1m: Large-scale stereo image dataset for 6d object pose
  estimation.
\newblock In {\em Proceedings of the IEEE/CVF International Conference on
  Computer Vision}, pages 10870--10879, 2021.

\bibitem{liu2020keypose}
Xingyu Liu, Rico Jonschkowski, Anelia Angelova, and Kurt Konolige.
\newblock Keypose: Multi-view 3d labeling and keypoint estimation for
  transparent objects.
\newblock In {\em Proceedings of the IEEE/CVF conference on computer vision and
  pattern recognition}, pages 11602--11610, 2020.

\bibitem{lopez2020project}
Adrian Lopez-Rodriguez, Benjamin Busam, and Krystian Mikolajczyk.
\newblock Project to adapt: Domain adaptation for depth completion from noisy
  and sparse sensor data.
\newblock In {\em Proceedings of the Asian Conference on Computer Vision},
  2020.

\bibitem{luo2020consistent}
Xuan Luo, Jia-Bin Huang, Richard Szeliski, Kevin Matzen, and Johannes Kopf.
\newblock Consistent video depth estimation.
\newblock {\em ACM Transactions on Graphics (Proceedings of ACM SIGGRAPH)},
  39(4):71--1, 2020.

\bibitem{mayer2018makes}
Nikolaus Mayer, Eddy Ilg, Philipp Fischer, Caner Hazirbas, Daniel Cremers,
  Alexey Dosovitskiy, and Thomas Brox.
\newblock What makes good synthetic training data for learning disparity and
  optical flow estimation?
\newblock {\em International Journal of Computer Vision}, 126(9):942--960,
  2018.

\bibitem{mayer2016large}
Nikolaus Mayer, Eddy Ilg, Philip Hausser, Philipp Fischer, Daniel Cremers,
  Alexey Dosovitskiy, and Thomas Brox.
\newblock A large dataset to train convolutional networks for disparity,
  optical flow, and scene flow estimation.
\newblock In {\em Proceedings of the IEEE conference on computer vision and
  pattern recognition}, pages 4040--4048, 2016.

\bibitem{Miangoleh2021Boosting}
S~Mahdi~H Miangoleh, Sebastian Dille, Long Mai, Sylvain Paris, and Yagiz Aksoy.
\newblock Boosting monocular depth estimation models to high-resolution via
  content-adaptive multi-resolution merging.
\newblock In {\em Proceedings of the IEEE/CVF Conference on Computer Vision and
  Pattern Recognition}, pages 9685--9694, 2021.

\bibitem{mildenhall2021nerf}
Ben Mildenhall, Pratul~P Srinivasan, Matthew Tancik, Jonathan~T Barron, Ravi
  Ramamoorthi, and Ren Ng.
\newblock Nerf: Representing scenes as neural radiance fields for view
  synthesis.
\newblock {\em Communications of the ACM}, 65(1):99--106, 2021.

\bibitem{kinectfusion}
Richard~A. Newcombe, Shahram Izadi, Otmar Hilliges, David Molyneaux, David Kim,
  Andrew~J. Davison, Pushmeet Kohi, Jamie Shotton, Steve Hodges, and Andrew
  Fitzgibbon.
\newblock Kinectfusion: Real-time dense surface mapping and tracking.
\newblock In {\em 2011 10th IEEE International Symposium on Mixed and Augmented
  Reality}, pages 127--136, 2011.

\bibitem{park2021nerfies}
Keunhong Park, Utkarsh Sinha, Jonathan~T Barron, Sofien Bouaziz, Dan~B Goldman,
  Steven~M Seitz, and Ricardo Martin-Brualla.
\newblock Nerfies: Deformable neural radiance fields.
\newblock In {\em Proceedings of the IEEE/CVF International Conference on
  Computer Vision}, pages 5865--5874, 2021.

\bibitem{qiu:2019a}
Simeng Qiu, Qiang Fu, Congli Wang, and Wolfgang Heidrich.
\newblock Polarization demosaicking for monochrome and color polarization focal
  plane arrays.
\newblock In Hans-Jörg Schulz, Matthias Teschner, and Michael Wimmer, editors,
  {\em Vision, Modeling and Visualization}. The Eurographics Association, 2019.

\bibitem{ranftl2021vision}
Ren{\'e} Ranftl, Alexey Bochkovskiy, and Vladlen Koltun.
\newblock Vision transformers for dense prediction.
\newblock In {\em Proceedings of the IEEE/CVF International Conference on
  Computer Vision}, pages 12179--12188, 2021.

\bibitem{Ranftl2022}
Ren\'{e} Ranftl, Katrin Lasinger, David Hafner, Konrad Schindler, and Vladlen
  Koltun.
\newblock Towards robust monocular depth estimation: Mixing datasets for
  zero-shot cross-dataset transfer.
\newblock {\em IEEE Transactions on Pattern Analysis and Machine Intelligence},
  44(3), 2022.

\bibitem{Reizenstein_2021_ICCV}
Jeremy Reizenstein, Roman Shapovalov, Philipp Henzler, Luca Sbordone, Patrick
  Labatut, and David Novotny.
\newblock Common objects in 3d: Large-scale learning and evaluation of
  real-life 3d category reconstruction.
\newblock In {\em Proceedings of the IEEE/CVF International Conference on
  Computer Vision (ICCV)}, pages 10901--10911, October 2021.

\bibitem{roessle2022dense}
Barbara Roessle, Jonathan~T Barron, Ben Mildenhall, Pratul~P Srinivasan, and
  Matthias Nie{\ss}ner.
\newblock Dense depth priors for neural radiance fields from sparse input
  views.
\newblock In {\em Proceedings of the IEEE/CVF Conference on Computer Vision and
  Pattern Recognition}, pages 12892--12901, 2022.

\bibitem{ruhkamp2021attention}
Patrick Ruhkamp, Daoyi Gao, Hanzhi Chen, Nassir Navab, and Benjamin Busam.
\newblock Attention meets geometry: Geometry guided spatial-temporal attention
  for consistent self-supervised monocular depth estimation.
\newblock In {\em IEEE International Conference on 3D Vision (3DV)}, December
  2021.

\bibitem{scharstein2002taxonomy}
Daniel Scharstein and Richard Szeliski.
\newblock A taxonomy and evaluation of dense two-frame stereo correspondence
  algorithms.
\newblock {\em International journal of computer vision}, 47(1):7--42, 2002.

\bibitem{silberman2012indoor}
Nathan Silberman, Derek Hoiem, Pushmeet Kohli, and Rob Fergus.
\newblock Indoor segmentation and support inference from rgbd images.
\newblock In {\em European conference on computer vision}, pages 746--760.
  Springer, 2012.

\bibitem{smith2018height}
William~AP Smith, Ravi Ramamoorthi, and Silvia Tozza.
\newblock Height-from-polarisation with unknown lighting or albedo.
\newblock {\em IEEE transactions on pattern analysis and machine intelligence},
  41(12):2875--2888, 2018.

\bibitem{son2016learning}
Kilho Son, Ming-Yu Liu, and Yuichi Taguchi.
\newblock Learning to remove multipath distortions in time-of-flight range
  images for a robotic arm setup.
\newblock In {\em 2016 IEEE International Conference on Robotics and Automation
  (ICRA)}, pages 3390--3397. IEEE, 2016.

\bibitem{spencer2020defeat}
Jaime Spencer, Richard Bowden, and Simon Hadfield.
\newblock Defeat-net: General monocular depth via simultaneous unsupervised
  representation learning.
\newblock In {\em Proceedings of the IEEE/CVF Conference on Computer Vision and
  Pattern Recognition}, pages 14402--14413, 2020.

\bibitem{replica19arxiv}
Julian Straub, Thomas Whelan, Lingni Ma, Yufan Chen, Erik Wijmans, Simon Green,
  Jakob~J. Engel, Raul Mur-Artal, Carl Ren, Shobhit Verma, Anton Clarkson,
  Mingfei Yan, Brian Budge, Yajie Yan, Xiaqing Pan, June Yon, Yuyang Zou,
  Kimberly Leon, Nigel Carter, Jesus Briales, Tyler Gillingham, Elias Mueggler,
  Luis Pesqueira, Manolis Savva, Dhruv Batra, Hauke~M. Strasdat, Renzo~De
  Nardi, Michael Goesele, Steven Lovegrove, and Richard Newcombe.
\newblock The {R}eplica dataset: A digital replica of indoor spaces.
\newblock {\em arXiv preprint arXiv:1906.05797}, 2019.

\bibitem{sturm2012benchmark}
J{\"u}rgen Sturm, Nikolas Engelhard, Felix Endres, Wolfram Burgard, and Daniel
  Cremers.
\newblock A benchmark for the evaluation of rgb-d slam systems.
\newblock In {\em 2012 IEEE/RSJ international conference on intelligent robots
  and systems}, pages 573--580. IEEE, 2012.

\bibitem{su2023opa}
Yongzhi Su, Yan Di, Guangyao Zhai, Fabian Manhardt, Jason Rambach, Benjamin
  Busam, Didier Stricker, and Federico Tombari.
\newblock Opa-3d: Occlusion-aware pixel-wise aggregation for monocular 3d
  object detection.
\newblock {\em IEEE Robotics and Automation Letters}, 2023.

\bibitem{sun2022direct}
Cheng Sun, Min Sun, and Hwann-Tzong Chen.
\newblock Direct voxel grid optimization: Super-fast convergence for radiance
  fields reconstruction.
\newblock In {\em Proceedings of the IEEE/CVF Conference on Computer Vision and
  Pattern Recognition}, pages 5459--5469, 2022.

\bibitem{CroMo}
Yannick Verdie, Jifei Song, Barnabé Mas, Busam Benjamin, Ales Leonardis, , and
  Steven McDonagh.
\newblock Cromo: Cross-modal learning for monocular depth estimation.
\newblock In {\em IEEE/CVF Conference on Computer Vision and Pattern
  Recognition (CVPR)}, 2022.

\bibitem{PhoCal}
Pengyuan Wang, HyunJun Jung, Yitong Li, Siyuan Shen, Rahul~Parthasarathy
  Srikanth, Loranzo Garattoni, Sven Meier, Nassir Navab, and Benjamin Busam.
\newblock Phocal: A multimodal dataset for category-level object pose
  estimation with photometrically challenging objects.
\newblock In {\em IEEE/CVF Conference on Computer Vision and Pattern
  Recognition (CVPR)}, 2022.

\bibitem{wang2021demograsp}
Pengyuan Wang, Fabian Manhardt, Luca Minciullo, Lorenzo Garattoni, Sven Meier,
  Nassir Navab, and Benjamin Busam.
\newblock Demograsp: Few-shot learning for robotic grasping with human
  demonstration.
\newblock In {\em 2021 IEEE/RSJ International Conference on Intelligent Robots
  and Systems (IROS)}, pages 5733--5740. IEEE, 2021.

\bibitem{wang2021nerf}
Zirui Wang, Shangzhe Wu, Weidi Xie, Min Chen, and Victor~Adrian Prisacariu.
\newblock Nerf--: Neural radiance fields without known camera parameters.
\newblock {\em arXiv preprint arXiv:2102.07064}, 2021.

\bibitem{watson2021temporal}
Jamie Watson, Oisin~Mac Aodha, Victor Prisacariu, Gabriel Brostow, and Michael
  Firman.
\newblock {The Temporal Opportunist: Self-Supervised Multi-Frame Monocular
  Depth}.
\newblock In {\em Computer Vision and Pattern Recognition (CVPR)}, 2021.

\bibitem{xiao2013sun3d}
Jianxiong Xiao, Andrew Owens, and Antonio Torralba.
\newblock Sun3d: A database of big spaces reconstructed using sfm and object
  labels.
\newblock In {\em Proceedings of the IEEE international conference on computer
  vision}, pages 1625--1632, 2013.

\bibitem{xie2016deep3d}
Junyuan Xie, Ross Girshick, and Ali Farhadi.
\newblock {Deep3d}: Fully automatic 2d-to-3d video conversion with deep
  convolutional neural networks.
\newblock In {\em European Conference on Computer Vision}, pages 842--857.
  Springer, 2016.

\bibitem{xie2022}
Yiheng Xie, Towaki Takikawa, Shunsuke Saito, Or Litany, Shiqin Yan, Numair
  Khan, Federico Tombari, James Tompkin, Vincent Sitzmann, and Srinath Sridhar.
\newblock Neural fields in visual computing and beyond.
\newblock {\em Computer Graphics Forum}, 2022.

\bibitem{yang2018lego}
Zhenheng Yang, Peng Wang, Yang Wang, Wei Xu, and Ram Nevatia.
\newblock Lego: Learning edge with geometry all at once by watching videos.
\newblock In {\em Proceedings of the IEEE conference on computer vision and
  pattern recognition}, pages 225--234, 2018.

\bibitem{Wei2021CVPR}
Wei Yin, Jianming Zhang, Oliver Wang, Simon Niklaus, Long Mai, Simon Chen, and
  Chunhua Shen.
\newblock Learning to recover 3d scene shape from a single image.
\newblock In {\em Proc. IEEE Conf. Comp. Vis. Patt. Recogn. (CVPR)}, 2021.

\bibitem{yu2017shape}
Ye Yu, Dizhong Zhu, and William~AP Smith.
\newblock Shape-from-polarisation: a nonlinear least squares approach.
\newblock In {\em Proceedings of the IEEE International Conference on Computer
  Vision Workshops}, pages 2969--2976, 2017.

\bibitem{zeng20163dmatch}
Andy Zeng, Shuran Song, Matthias Nie{\ss}ner, Matthew Fisher, Jianxiong Xiao,
  and Thomas Funkhouser.
\newblock 3dmatch: Learning local geometric descriptors from rgb-d
  reconstructions.
\newblock In {\em CVPR}, 2017.

\bibitem{zhai2022monograspnet}
Guangyao Zhai, Dianye Huang, Shun-Cheng Wu, HyunJun Jung, Yan Di, Fabian
  Manhardt, Federico Tombari, Nassir Navab, and Benjamin Busam.
\newblock Monograspnet: 6-dof grasping with a single rgb image.
\newblock In {\em IEEE International Conference on Robotics and Automation}.
  IEEE, 2023.

\bibitem{da2dataset}
Guangyao Zhai, Yu Zheng, Ziwei Xu, Xin Kong, Yong Liu, Benjamin Busam, Yi Ren,
  Nassir Navab, and Zhengyou Zhang.
\newblock Da$^2$ dataset: Toward dexterity-aware dual-arm grasping.
\newblock {\em IEEE Robotics and Automation Letters}, 7(4):8941--8948, 2022.

\bibitem{Zhou2018}
Qian-Yi Zhou, Jaesik Park, and Vladlen Koltun.
\newblock {Open3D}: {A} modern library for {3D} data processing.
\newblock {\em arXiv:1801.09847}, 2018.

\bibitem{zhu2019depth}
Dizhong Zhu and William~AP Smith.
\newblock Depth from a polarisation + rgb stereo pair.
\newblock In {\em Proceedings of the IEEE/CVF Conference on Computer Vision and
  Pattern Recognition}, pages 7586--7595, 2019.

\end{thebibliography}
}

\end{document}

\title{On the Importance of Accurate Geometry Data for Dense 3D Vision Tasks \\ -- \\ Supplementary Material}

% \and
\author{
\hspace{-20pt}
HyunJun Jung$^{\ast 1}$,
Patrick Ruhkamp$^{\ast 1,2}$,
Guangyao Zhai$^{1}$,
Nikolas Brasch$^{1}$,
Yitong Li$^{1}$,\\
Yannick Verdie$^{1,3}$,
Jifei Song$^{3}$,
Yiren Zhou$^{3}$,
Anil Armagan$^{3}$,
Slobodan Ilic$^{1,4}$,\\
Ales Leonardis$^{3}$,
Nassir Navab$^{1}$,
Benjamin Busam$^{1,2}$
% HyunJun Jung$^{\ast 1}$\and
% Patrick Ruhkamp$^{\ast 1,2}$\and
% Guangyao Zhai$^{1}$\and
% Nikolas Brasch$^{1}$\and
% Yitong Li$^{1}$\and
% Yannick Verdie$^{1,3}$\and
% Jifei Song$^{3}$\and
% Yiren Zhou$^{3}$\and
% Anil Armagan$^{3}$\and
% Slobodan Ilic$^{1,4}$\and
% Ales Leonardis$^{3}$\and
% Nassir Navab$^{1}$\and
% Benjamin Busam$^{1,2}$
\\
\\
\small
$^1$ Technical University of Munich,
$^2$ 3Dwe.ai,
$^3$  Huawei Noah's Ark Lab,
$^4$  Siemens AG,
$^*$ Equal Contribution
\\ \footnotesize{\fontfamily{qcr}\selectfont
hyunjun.jung@tum.de, p.ruhkamp@tum.de, guangyao.zhai@tum.de, b.busam@tum.de
}
}
\maketitle

\noindent
\section{Dense 3D Vision Tasks}
\subsection{Monocular Depth Estimation}
\label{sec:extra_evaluation}
Following the results on monocular depth estimation in the main paper, we describe the implementation details of the training, show additional results on different scenes and provide additional metrics on different test scenes. 

\paragraph{Implementation Details}
For all our depth estimation experiments, we use PyTorch~\cite{paszke2017automatic} and train for 20 epochs for comparability using Adam~\cite{kingma2014adam}.
Monocular approaches are trained with a batch size of 12 on one NVIDIA RTX-3090 GPU.
We chose $\lambda_{\text{s}}=10^{-3}$ and sample $S$ with $T=10$ frames offset due to small relative camera movement between frames and the high frame rate.
The RGB inputs are scaled to $480 \times 320$ for supervised training and to $320 \times 160$ for self-supervised training, respectively. 
The depth network regresses dense depth predictions on four pyramid levels, each with half the resolution of the previous.
Pose network and augmentations follow~\cite{monodepth2}.
We choose an initial learning rate of $1 \times 10^{-4}$ for 15 epochs, which we decrease to $1 \times 10^{-5}$ after 15 epochs in the self-supervised setting.
For the supervised case, we start with a learning rate of $1 \times 10^{-3}$, which we decrease every five epochs by a factor of ten.

\begin{table*}[!ht]
\centering
\caption{\textbf{Depth prediction comparison when training with different modalities and tested on different unseen scenes and seen scenes. }(Top) Evaluation against GT of depth predictions on the test set with dense supervision from different depth modalities. (Bottom) Predictions evaluated on respective modality. 
Error is reported as Sq.Rel. and RMSE in mm.
}
\begin{adjustbox}{}
\footnotesize
\resizebox{1.0\textwidth}{!}{
\begin{tabular}{ll|cc|cccc|cccccc}
\shline
& Mask&\multicolumn{2}{c|}{Full Scene} & \multicolumn{2}{c}{Background} & \multicolumn{2}{c|}{All Objects}  & \multicolumn{2}{c}{Textured} & \multicolumn{2}{c}{Reflective} & \multicolumn{2}{c}{Transparent} \\
\hline
&Metric&Sq.Rel. & RMSE&Sq.Rel. & RMSE&Sq.Rel. & RMSE&Sq.Rel. & RMSE&Sq.Rel. & RMSE&Sq.Rel. & RMSE \\
\shline
\multirow{3}{*}{\rotatebox[origin=c]{-90}{Test 1}}     & I-ToF & 
24.78  &  148.09
 & 
22.25  &  151.07
 & 
29.62  &  123.19
 & 
16.47  &  99.08
 & 
102.79  &  214.60
 & 
44.29  &  134.44
\\
                                & D-ToF & 
24.23  &  151.72
 & 
23.74  &  159.28
 & 
22.85  &  110.88
 & 
16.22  &  101.12
 & 
57.14  &  148.61
 & 
30.23  &  107.23
\\
                                & Active Stereo & 
32.15  &  173.72
 & 
33.84  &  184.16
 & 
22.23  &  116.57
 & 
19.55 &  114.07
 & 
64.27  &  167.71
 & 
12.92  &  69.49
\\
\shline\\
\shline

\multirow{3}{*}{\rotatebox[origin=c]{-90}{Test 2}}     & I-ToF & 
27.42  &  123.79
 & 
22.66  &  116.86
 & 
39.85  &  139.67
 & 
48.66  &  144.92
 & 
16.15  &  99.44
 & 
25.15  &  122.25
\\
                                & D-ToF & 
23.00  &  115.40
 & 
21.18  &  113.27
 & 
27.89  &  119.59
 & 
30.00  &  112.92
 & 
15.81  &  90.89
 & 
23.73  &  117.72
\\
                                & Active Stereo & 
25.94  &  124.17
 & 
25.50  &  126.28
 & 
27.18  &  117.04
 & 
32.81  &  121.24
 & 
16.40  &  101.86
 & 
15.73  &  95.27
\\              
\shline\\
\shline

\multirow{3}{*}{\rotatebox[origin=c]{-90}{Test 3}}     & I-ToF & 
36.82  &  152.51
 & 
35.92  &  153.26
 & 
38.75  &  147.14
 & 
34.09  &  127.51
 & 
20.21  &  110.85
 & 
55.09  &  183.14
\\
                                & D-ToF & 

32.99  &  145.50
 & 
35.64  &  153.07
 & 
25.90  &  120.35
 & 
19.92  &  96.01
 & 
21.59  &  105.41
 & 
37.26  &  149.66
\\
                                & Active Stereo & 
31.63  &  141.77
 & 
35.24  &  151.37
 & 
22.44  &  110.42
 & 
23.47  &  106.63
 & 
14.49  &  94.51
 & 
21.21  &  109.53
\\                              
\shline\\
\shline

\multirow{3}{*}{\rotatebox[origin=c]{-90}{T. Seen}}     & I-ToF & 
9.87  &  77.99
 & 
4.62  &  57.10
 & 
33.91  &  133.46
 & 
6.18  &  60.48
 & 
35.65  &  119.76
 & 
91.30  &  224.27
\\
                                & D-ToF & 

15.43  &  93.31
 & 
11.62  &  79.89
 & 
31.12  &  123.97
 & 
4.40  &  51.91
 & 
17.42  &  82.29
 & 
89.19  &  212.55
\\
                                & Active Stereo & 
9.43  &  88.30
 & 
9.28  &  88.24
 & 
9.11  &  75.21
 & 
6.32  &  65.54
 & 
12.98  &  65.73
 & 
16.62  &  98.75
\\
\shline\\
\\
\\
\multicolumn{6}{l}{Tested on Modality:}
\\
\shline
\multirow{4}{*}{\rotatebox[origin=c]{-90}{Test Seen}}     & I-ToF & 
8.34  &  52.29
 & 
8.57  &  50.00
 & 

7.01  &  58.85
&
3.80  &  43.44
 & 
23.28  &  95.38
 & 
13.69  &  65.41
\\
                                & D-ToF & 

8.05  &  50.43
 & 
6.82  &  45.50
 & 
 
13.52  &  66.34
&
9.00  &  54.15
 & 
30.91  &  87.71
 & 
27.92  &  87.32
\\
                                & Active Stereo & 
39.25  &  101.76
 & 
40.87  &  102.29
 & 
 
30.32  &  90.00
&
32.24  &  90.49
 & 
23.36  &  72.21
 & 
37.25  &  101.23
\\
                                & GT & 
1.12  &  28.81
 & 
0.71  &  24.41
 & 
 
2.65  &  40.41
&
1.83  &  34.89
 & 
2.16  &  29.55
 & 
5.02  &  52.43
\\
\shline

\end{tabular}
}
\end{adjustbox}
\label{tab:depth_supervision_results_additional}
\end{table*}

\subsubsection{Quantitative evaluation}
\paragraph{Test scenes. }
Table~\ref{tab:depth_supervision_results_additional} summarizes the extensive quantitative evaluation of the supervised training with different depth modalities as supervision signal for different test scenes. Test scene 1 has a similar background compared to the training scenes and includes additional unseen objects. The scene is also observed from viewing angles that differ significantly from the training data. The background in test scene 2 is only partly observed in the training data and it includes mostly unseen objects. Test scene 3 is similar to test scene 2, but with a modified object layout and difficult lighting in the background from an additional bright light source above the scene. The additional test set with (partly) seen scenes is an additional test split which includes the first 10 frames of each training sequence. Please note that these frames have not been used during training. Here, we first test all predictions against the rendered ground truth (Top) and additionally on each individual respective modality (Bottom) to highlight the overfitting issue of invalid ground truth from each modality.
The results suggest that overall the supervision with accurate rendered ground truth achieves to generalize best for (mostly) unknown scenes. It is noticeable, that the active stereo achieves to produce good predictions for transparent objects and also performs well for reflective ones. The I-ToF and D-ToF predictions suffer from incorrect ground truth values for such objects.

\paragraph{Overfitting on (partly) seen scenes. }
The (partly) seen scene shows generally lower overall errors for all modalities as compared to the (mostly) unseen test scenes 1,2, and 3. Again, the active stereo can provide decent depth supervision for reflective and transparent objects, where the ToF sensors cannot provide valid depth. The prediction of the background of the scene performs worst for the active stereo, as the textureless wall is still problematic for the sensor.

When testing on the respective modality itself, the overfitting issue due to incorrect depth values of the sensor becomes apparent. It can be noticed, that for objects where the respective sensor cannot yield accurate depth values (e.g. transparent objects for I-ToF or reflective objects for D-ToF), the errors are significantly lower, indicating overfitting to the specific sensor modality.

\subsubsection{Qualitative predictions}
\begin{figure*}[!p]
 \centering
    \includegraphics[width=\linewidth]{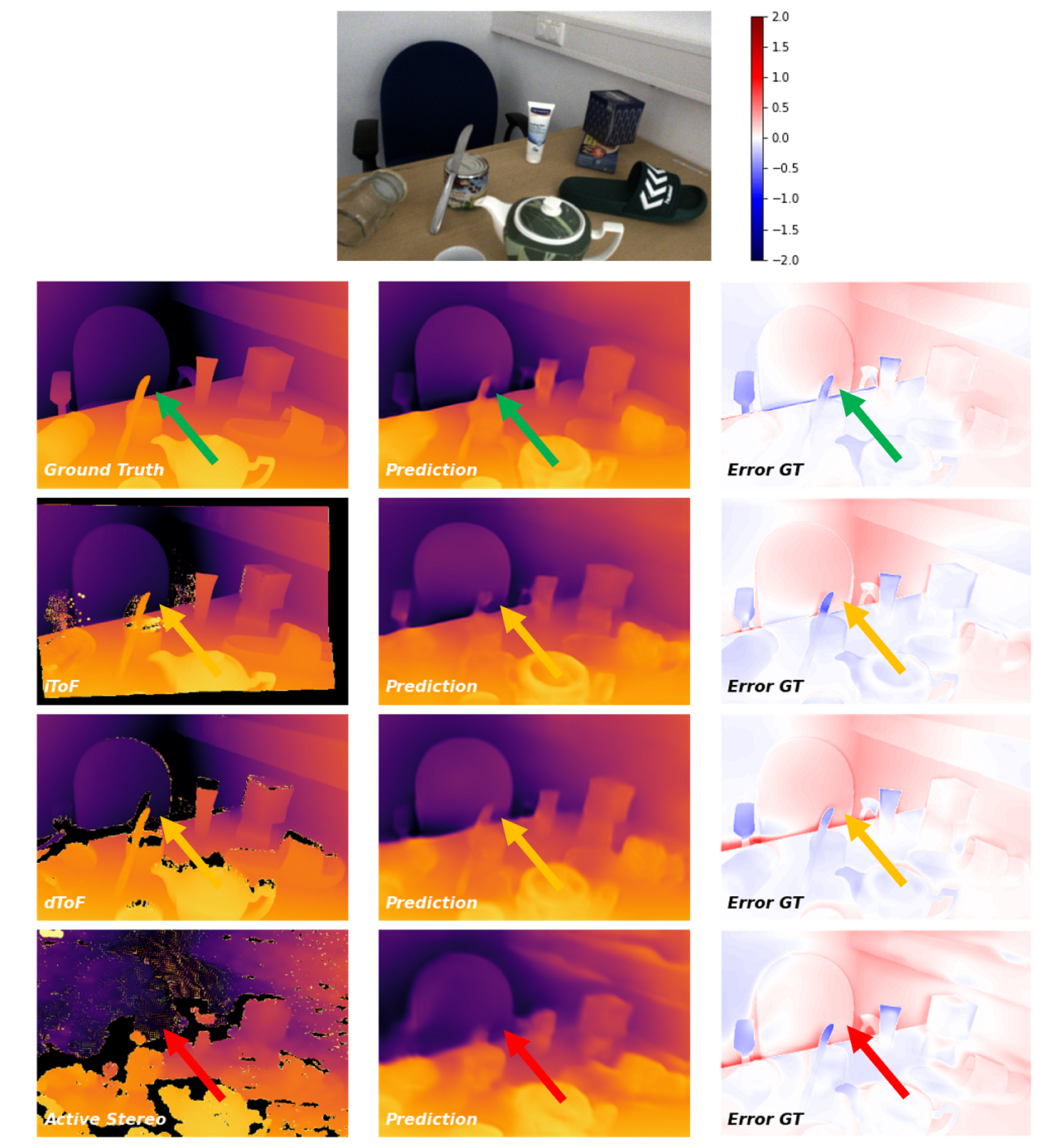}
    \caption{Qualitative evaluation on test scene 1. Each depth modality, the network prediction when trained with supervision of each modality, and the error, are shown as qualitative evaluation.}
    \label{fig:qual_scene12_1}
\end{figure*}

\begin{figure*}[!p]
 \centering
    \includegraphics[width=\linewidth]{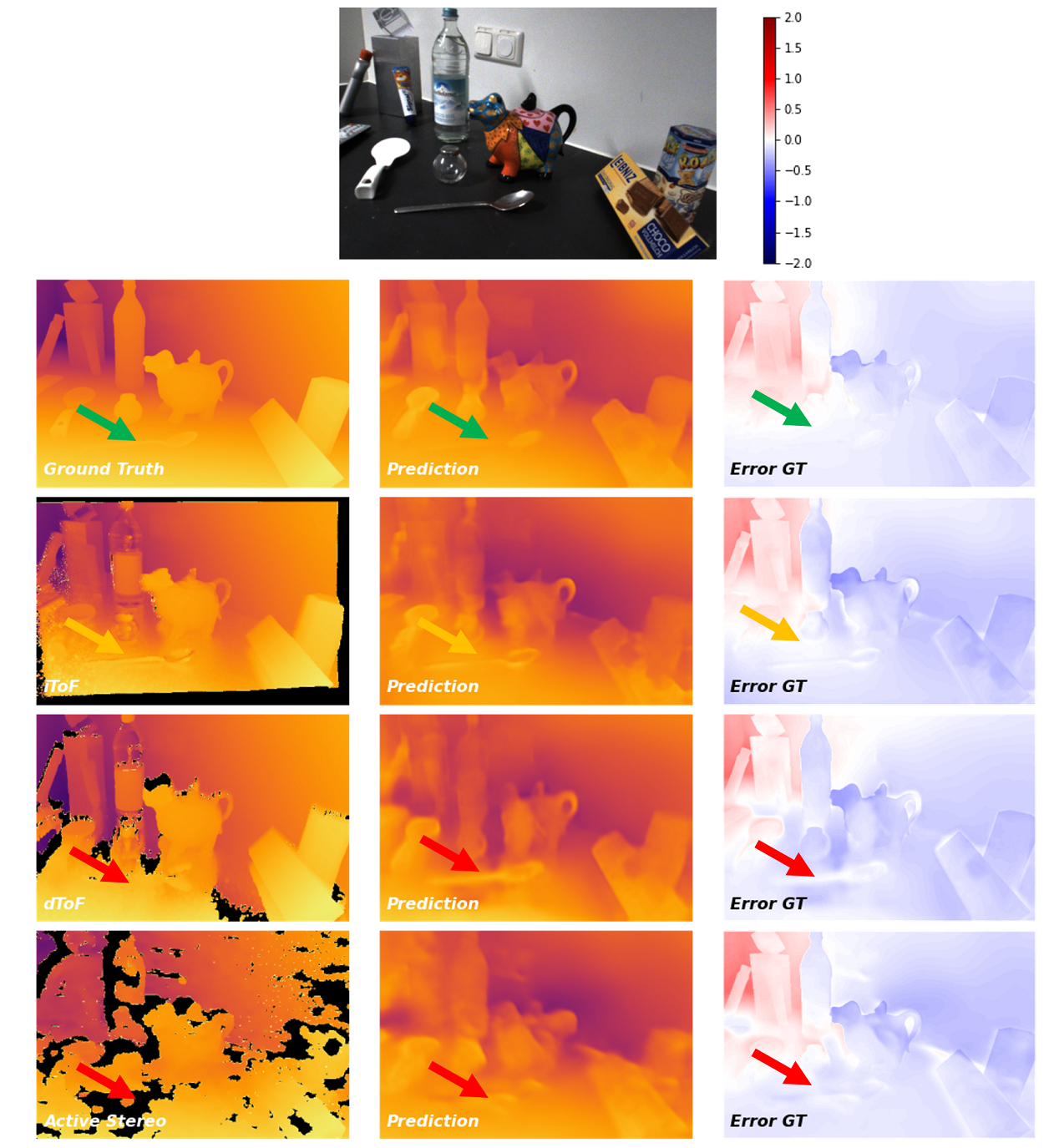}
    \caption{Qualitative evaluation on test scene 2. Each depth modality, the network prediction when trained with supervision of each modality, and the error, are shown as qualitative evaluation.}
    \label{fig:qual_scene13_1}
\end{figure*}

\begin{figure*}[!p]
 \centering
    \includegraphics[width=\linewidth]{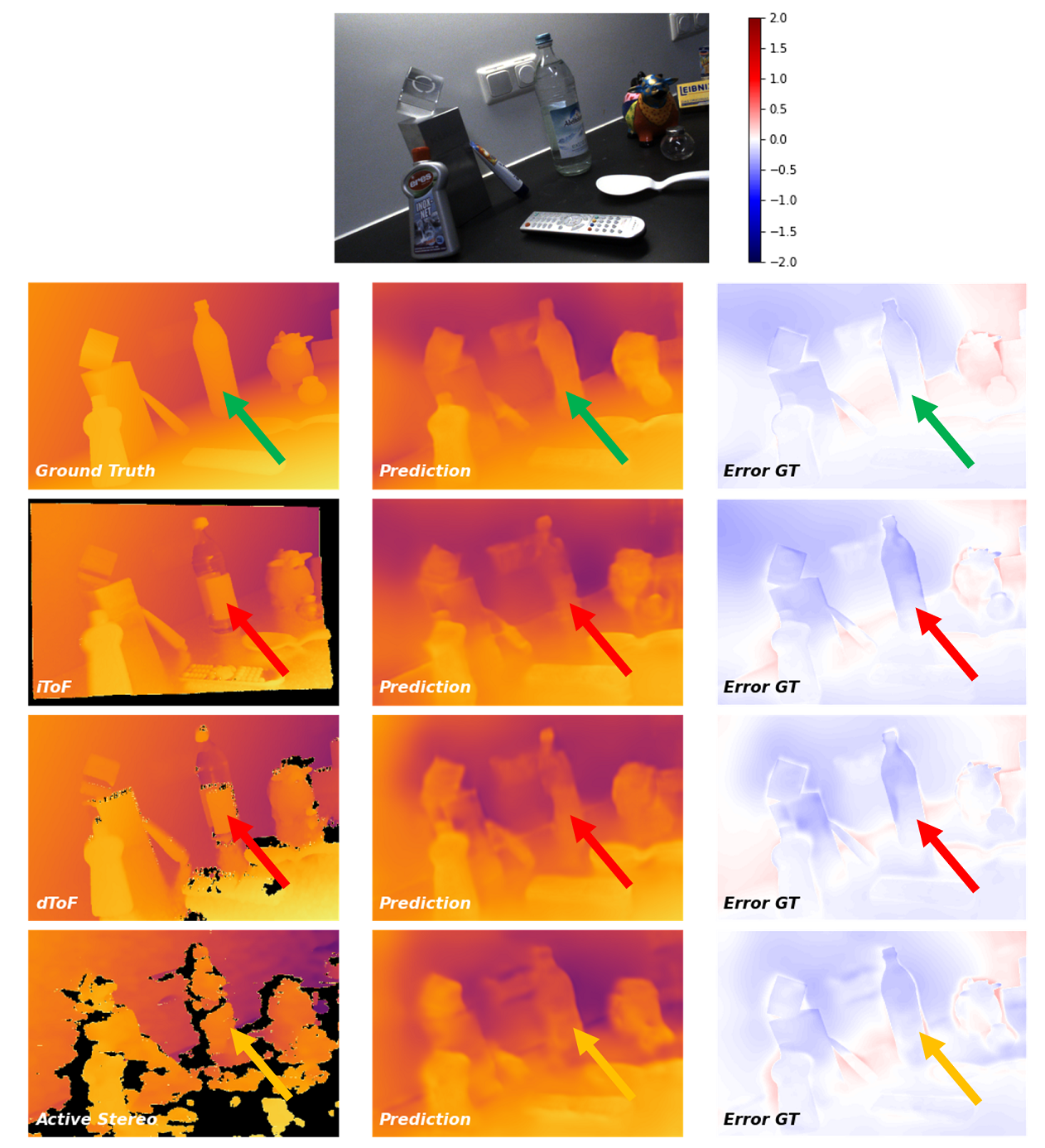}
    \caption{Qualitative evaluation on test scene 3. Each depth modality, the network prediction when trained with supervision of each modality, and the error, are shown as qualitative evaluation.}
    \label{fig:qual_scene14_1}
\end{figure*}

Figures~\ref{fig:qual_scene12_1},~\ref{fig:qual_scene13_1} and~\ref{fig:qual_scene14_1} show predictions on exemplary frames of the test scenes 1, 2 and 3, together with the different sensor modalities and the error plot of the prediction compared against the ground truth. The training with rendered ground truth generally performs best. Both ToF sensors show incorrect depth values for reflective or transparent objects which also translates to incorrect predictions in these areas (compare Fig.~\ref{fig:qual_scene12_1}. The predictions when training with active stereo as supervision are more blurry and show less distinct edges at depth boundaries when compared to other modalities, which may arise from many depth pixels being invalidated by the sensors around such boundaries (compare Fig.~\ref{fig:qual_scene13_1}). The very challenging test scene 3 with bright lighting and many unseen objects is difficult to predict for all training setups (compare Fig.~\ref{fig:qual_scene14_1}. We can see similar artifacts as described above. Additionally, the unseen trophy object with partly reflective and partly transparent material shows large errors for the sensor inputs as well as for its predictions. The desk surface is also incorrectly captured by the D-ToF sensors due to large reflections and MPI from the background.

\subsection{Implicit Reconstruction}
\paragraph{Implementation Details}
As mentioned in the main paper, we follow NeRF~\cite{mildenhall2021nerf} and build upon the work of~\cite{roessle2022dense} without a depth completion network, but leverage the respective sensor depth with a scale-invariant depth loss $\mathcal{L}_{\text{D}}$. We use images with a resolution of $640 \times 480$ and process batches of 1024 rays. We set $\lambda_{\text{D}}$ to 0.1 and the learning rate to 0.0005 and optimize for 100k iterations with Adam optimizer~\cite{kingma2014adam}. 

\subsection{Camera Pose Estimation}
\begin{figure*}[h!]
 \centering
    \includegraphics[width=1.0\textwidth]{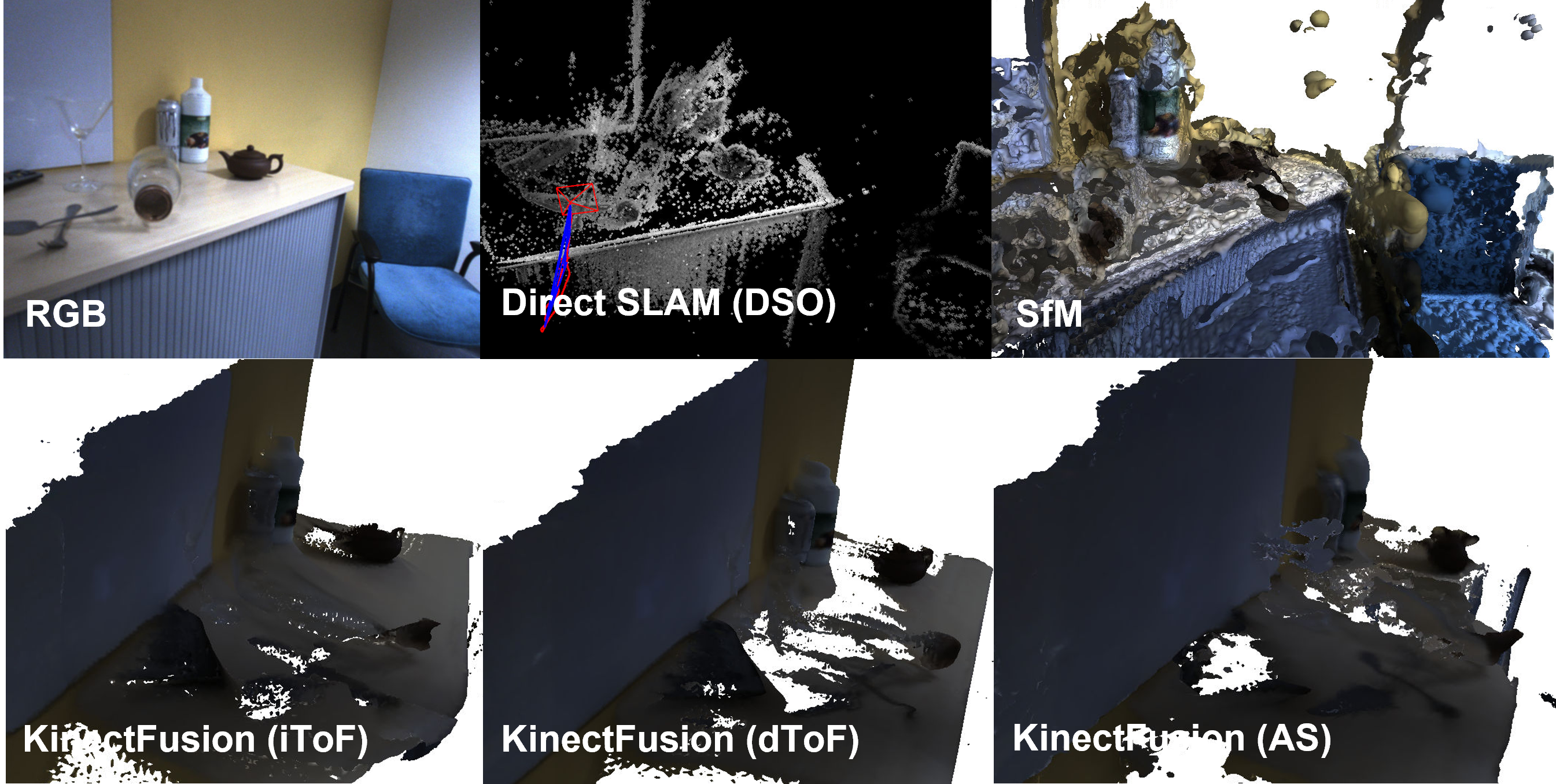}
    \caption{Qualitative reconstruction results from SLAM and SfM.}
    \label{fig:recon_comp}
\end{figure*}

The analysis above focuses on dense monocular depth estimation and novel view synthesis as recent and important approaches - for which pixelwise prediction and evaluation are crucial. We add results for direct SLAM (DSO)~\cite{engel2017direct}, KinectFusion~\cite{kinectfusion} with different depth modalities, and COLMAP SfM~\cite{schoenberger2016sfm} in Fig.~\ref{fig:recon_comp}.

Tab.~\ref{tab:pose} summarizes the relative pose error for different approaches (cf. Fig~\ref{fig:recon_comp}). Note the pose accuracy results for KinectFusion~\cite{kinectfusion} align with the depth results from Tab.2 in the main paper.

\begin{table}[!t]
\centering
\caption{Relative Pose Error of SLAM and SfM.}
\resizebox{1.0\columnwidth}{!}{
    \centering 
    \begin{tabular}{l|c|ccc|c} 
    \shline
    Error & Direct (DSO) & Dense dToF & Dense iToF & Dense AS & SfM \\ 
    \shline
    rot [deg] & 0.22 & 0.18 & 0.51 & 0.56 & 10.76 \\
    trans [cm] & 0.27 & 0.31 & 0.68 & 0.62 & 2.86 \\
    \shline
    \end{tabular}}
    \label{tab:pose}
\end{table}

\clearpage

\section{Dataset}
\subsection{Detailed Dataset Description}
\label{sec:detailed_dataset_description}

Sec.~3 of the main paper mentioned that our dataset uses multiple images/depth sensors to collect the dataset with highly accurate annotations of the scene using the robotic arm in a synchronized manner. This section shows the detailed description of data we include in our dataset.

\subsubsection{Polarization Camera}
\label{subsec:polarization_camera}

\begin{figure*}[!htbp]
 \centering
    \includegraphics[width=\linewidth]{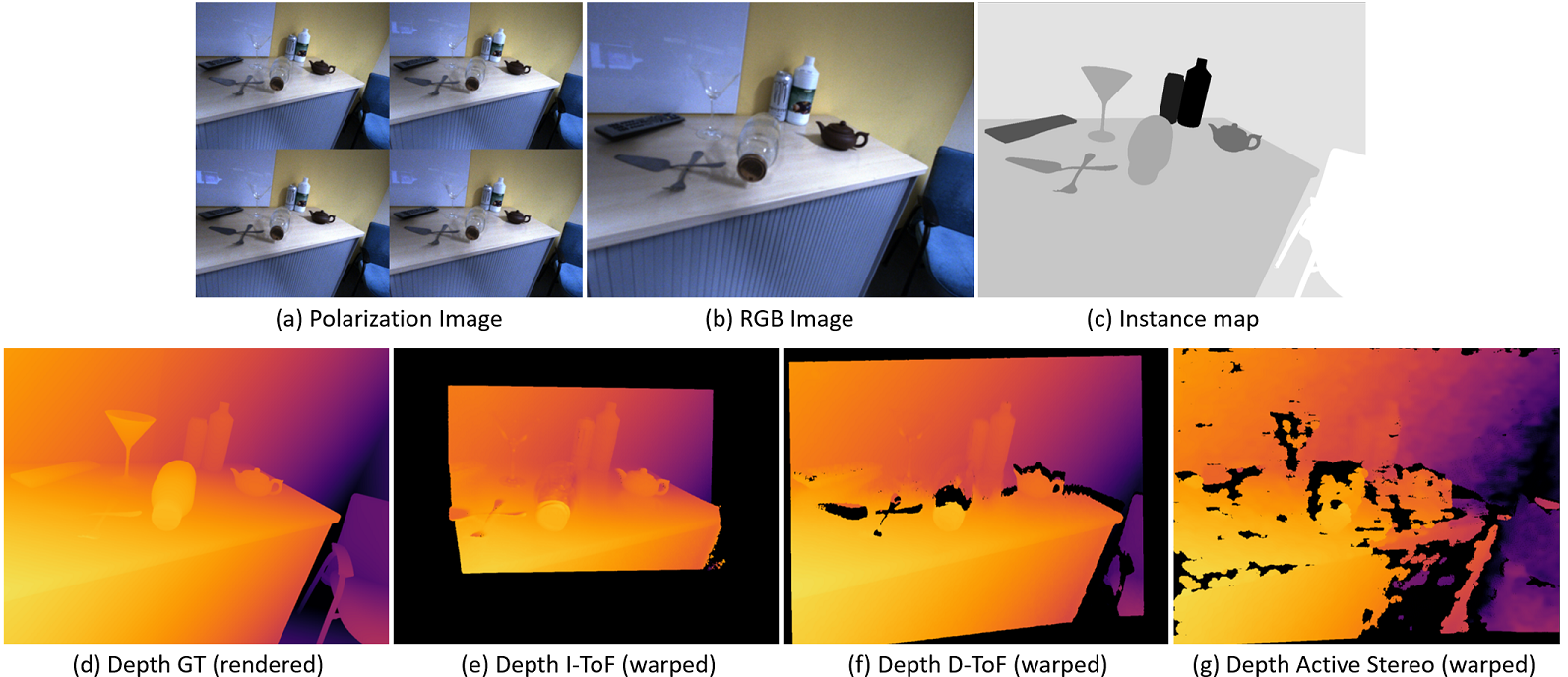}
    \caption{Example of the images included for the polarization camera input (top) together with instance label map and depth estimates warped onto the same coordinate reference frame.}
    \label{fig:polarization_included}
\end{figure*}

Fig.~\ref{fig:polarization_included} shows examples of images included for the polarization camera. As mentioned in Sec.~2 of the main paper, a polarization camera provides images with different polarization angles, which can extract cues like the surface normal by using the physical property of object material in the scene. The polarization camera we used in our dataset (See Sec.~3 in the main paper) provides polarized images at 4 different angles (0, 90, 180 270 degrees) which are saved in a single 2x2 image (Fig.~\ref{fig:polarization_included}, (a)). A regular RGB image is obtained by averaging the 4 images (Fig.~\ref{fig:polarization_included}, (b)). To showcase the results of the depth map trained with different depth cameras, we include warped depth images from each depth camera into the polarization camera coordinates using the extrinsic between the two cameras and its depth image (Fig.~\ref{fig:polarization_included}, (d-g)). These can be additionally used for RGBD-based depth completion research. On top of that, we include extra information, such as instance map (Fig.~\ref{fig:polarization_included}, (c)) to help train or validate pipelines for categorical level tasks, accurate 6d pose of the camera as the 4x4 matrix obtained from the robotic arm, extrinsic transformation between cameras as 4x4 matrices and camera intrinsics as 3x3 matrix.

\subsubsection{D-ToF Camera}
\label{subsec:dtof_camera}

\begin{figure*}[!htbp]
 \centering
    \includegraphics[width=\linewidth]{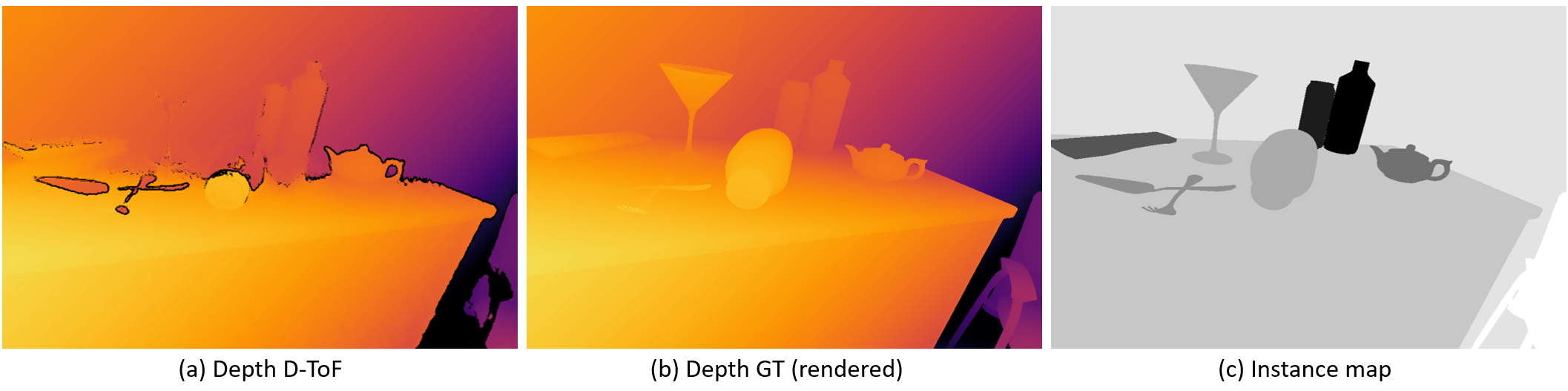}
    \caption{Example of the images included for the D-ToF camera: its depth map (left), ground truth depth (centre) and an object instance label map (right).}
    \label{fig:dtof_included}
\end{figure*}

Fig.~\ref{fig:dtof_included} shows an example of images included for the D-ToF camera. Direct ToF (D-ToF) camera senses the depth information of its surrounding by emitting an infrared signal and measuring the difference in time between the emitted and received signal. The quality of this modality highly depends on the reflection of the signal. It often suffers from specific physical noise such as Multi-Path-Interference (MPI) or strong material dependent artefacts (Fig.~\ref{fig:dtof_mpi}). For the D-ToF camera, we provide the depth map from the camera (Fig.~\ref{fig:dtof_included}, (a)) as well as its rendered ground truth depth map (Fig.~\ref{fig:dtof_included}, (b))  such that one can also research on D-ToF refinement pipelines to reduce such errors. As in the polarization camera, we include extra information such as instance label map (Fig.~\ref{fig:dtof_included}, (c)), camera pose, intrinsic and extrinsics of the camera as well.

\subsubsection{I-ToF Camera}
\label{subsec:itof_camera}

\begin{figure*}[!htbp]
 \centering
    \includegraphics[width=\linewidth]{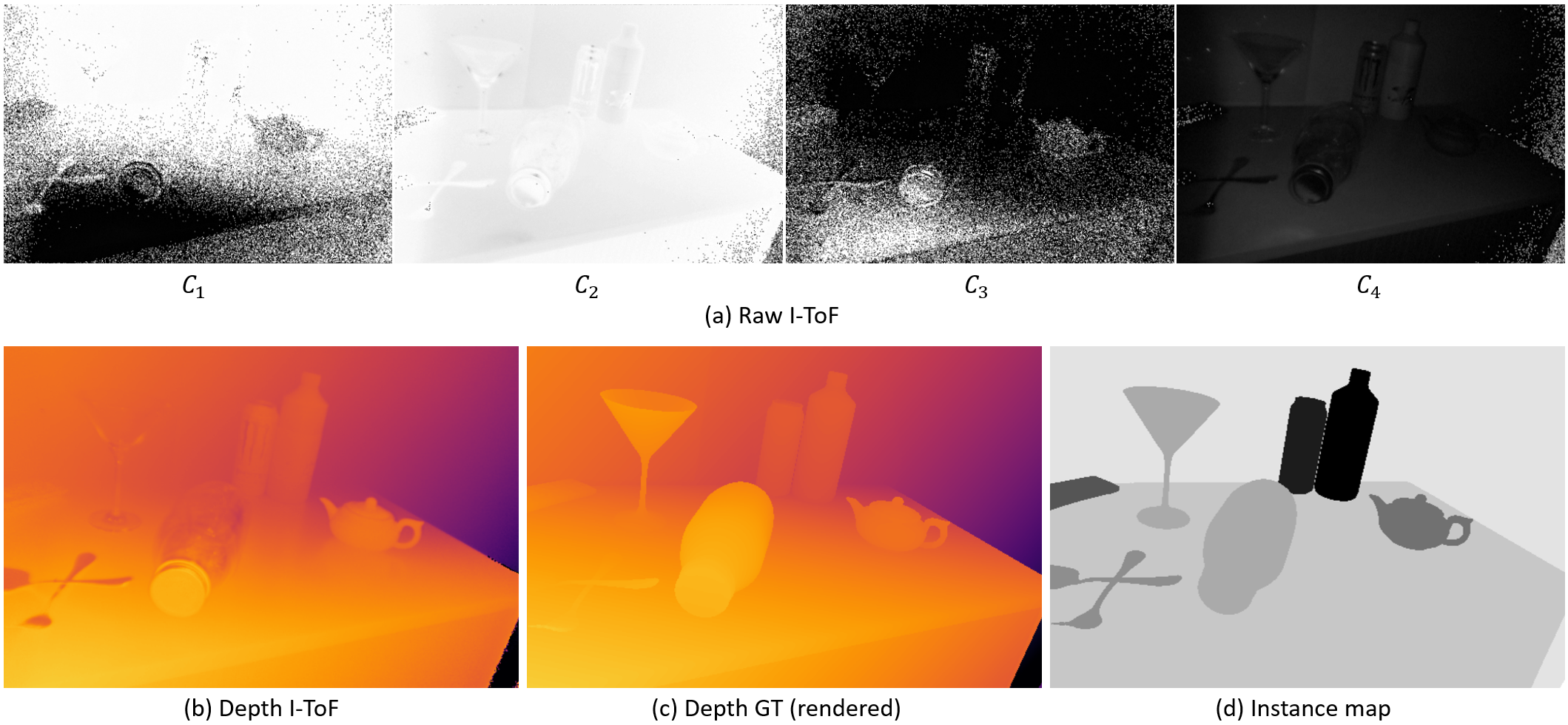}
    \caption{Example of the images included for the I-ToF camera.}
    \label{fig:itof_included}
\end{figure*}

Fig.~\ref{fig:itof_included} shows image examples for the I-ToF camera. Indirect ToF (I-ToF) cameras sense the depth information of their surrounding by emitting a frequency modulated signal and measuring the return signal. Unlike Direct ToF (D-ToF), I-ToF cameras do not calculate the time difference to infer the depth. Instead, the camera correlates the returning signal with phase-shifted emitting signals to generate 4 different measurements, called correlation images. These are measured as sinus functions of distance ($\left( \sin(d),\cos(d),-\sin(d),-\cos(d) \right) = \left( c_{1}, c_{2}, c_{3}, c_{4} \right)$ in Fig.~\ref{fig:itof_included}, (a)). Either arc-tangent formula or convolutional neural networks can be used to extract depth information from the correlation images. As I-ToF modality also relies on the reflection of the signal like in D-ToF, it suffers from similar artefacts, such as MPI and material dependent artefacts (compare qualitative results of the test scenes in Figs.~\ref{fig:qual_scene12_1}, ~\ref{fig:qual_scene13_1} and ~\ref{fig:qual_scene14_1}). Here, we provide raw correlation images and depth map from the camera (see Fig.~\ref{fig:itof_included}, (a,b)) as well as its rendered ground truth depth (Fig.~\ref{fig:itof_included}, (c)) such that one can train I-ToF depth improvement pipelines either from raw signal or from I-ToF depth itself. As the other cameras, extras such as instance map (Fig.~\ref{fig:itof_included}, (d)), camera pose, intrinsic and extrinsics are included.

\subsubsection{Active Stereo Camera}
\label{subsec:active_stereo_camera}

\begin{figure*}[!htbp]
 \centering
    \includegraphics[width=\linewidth]{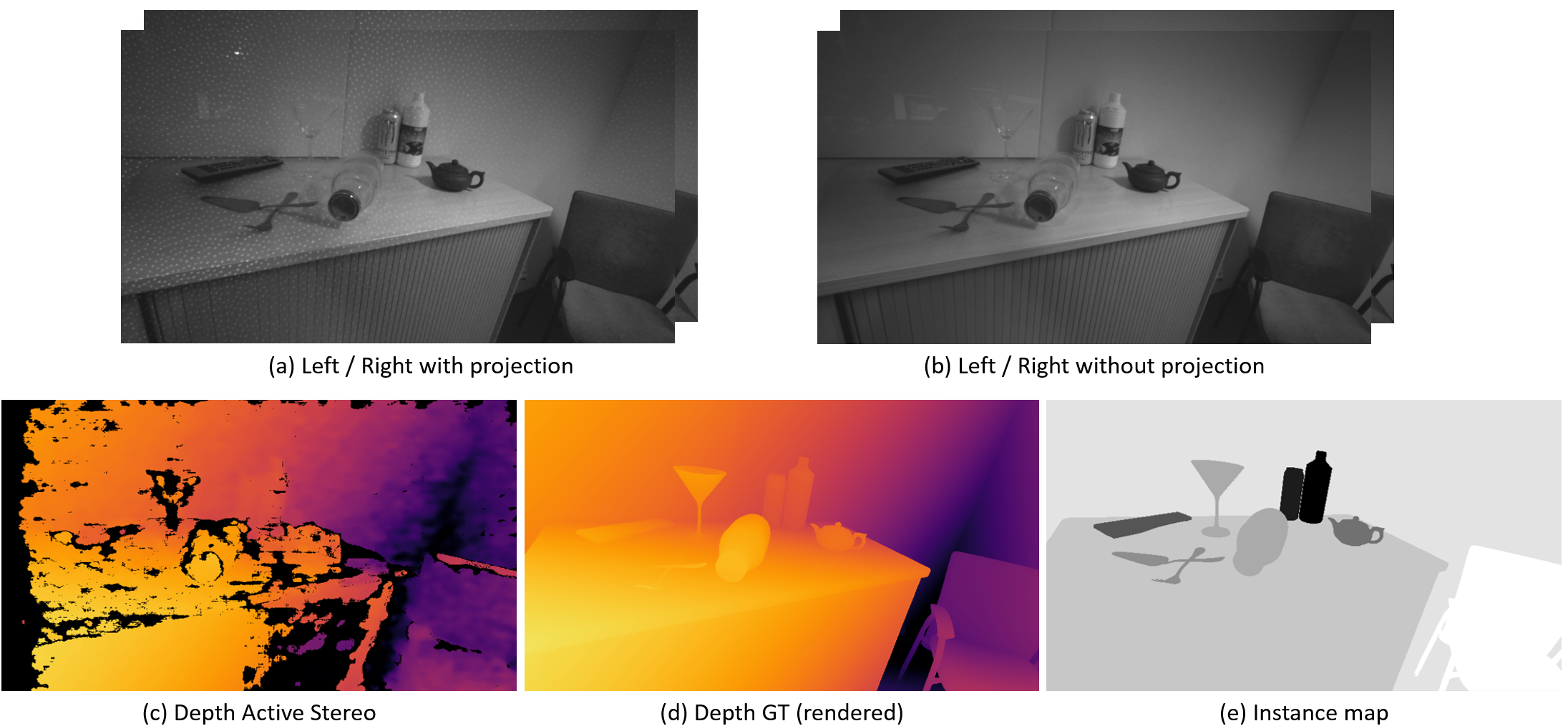}
    \caption{Example of the images included for the Active Stereo camera.}
    \label{fig:d435_included}
\end{figure*}

Fig.~\ref{fig:d435_included} shows the examples of images included for the Active Stereo camera. Stereo depth estimation infers depth using and photometric consistency and geometrical constraints from epipolar geometry and triangulates the depth map from the disparity between left and right cameras. As the disparity is calculated via matching on the image itself, the stereo based depth estimation methods suffers less from the specific material, but they suffer from other aspects such as stereo occlusion and large texture-less regions. Active projection (Active Stereo) is used to overcome this issue. We provide both, active and passive stereo left / right images (Fig.~\ref{fig:d435_included}, (a),(b)) and raw depth from the camera (active,  Fig.~\ref{fig:d435_included}, (c)) as well as the rendered ground truth (Fig.~\ref{fig:d435_included}, (d)). This allows to use our dataset to improve stereo methods from either passive or active stereo and also depth refinement pipelines. Similar to the other cameras, extras such as instance map (Fig.~\ref{fig:d435_included}, (e)), camera pose, intrinsic and extrinsics are included.

\subsection{Error Analysis on Different Modality}
\label{sec:Error_analysis}
In this section, we show specific errors on each depth modality to illustrate the implication of the depth quality when the given modality is used as the ground truth, as well as advantage of using our rendered depth as the ground truth.

\subsubsection{D-ToF Camera}
\label{subsec:dtof_camera_error}
\begin{figure*}[!htbp]
 \centering
    \includegraphics[width=\linewidth]{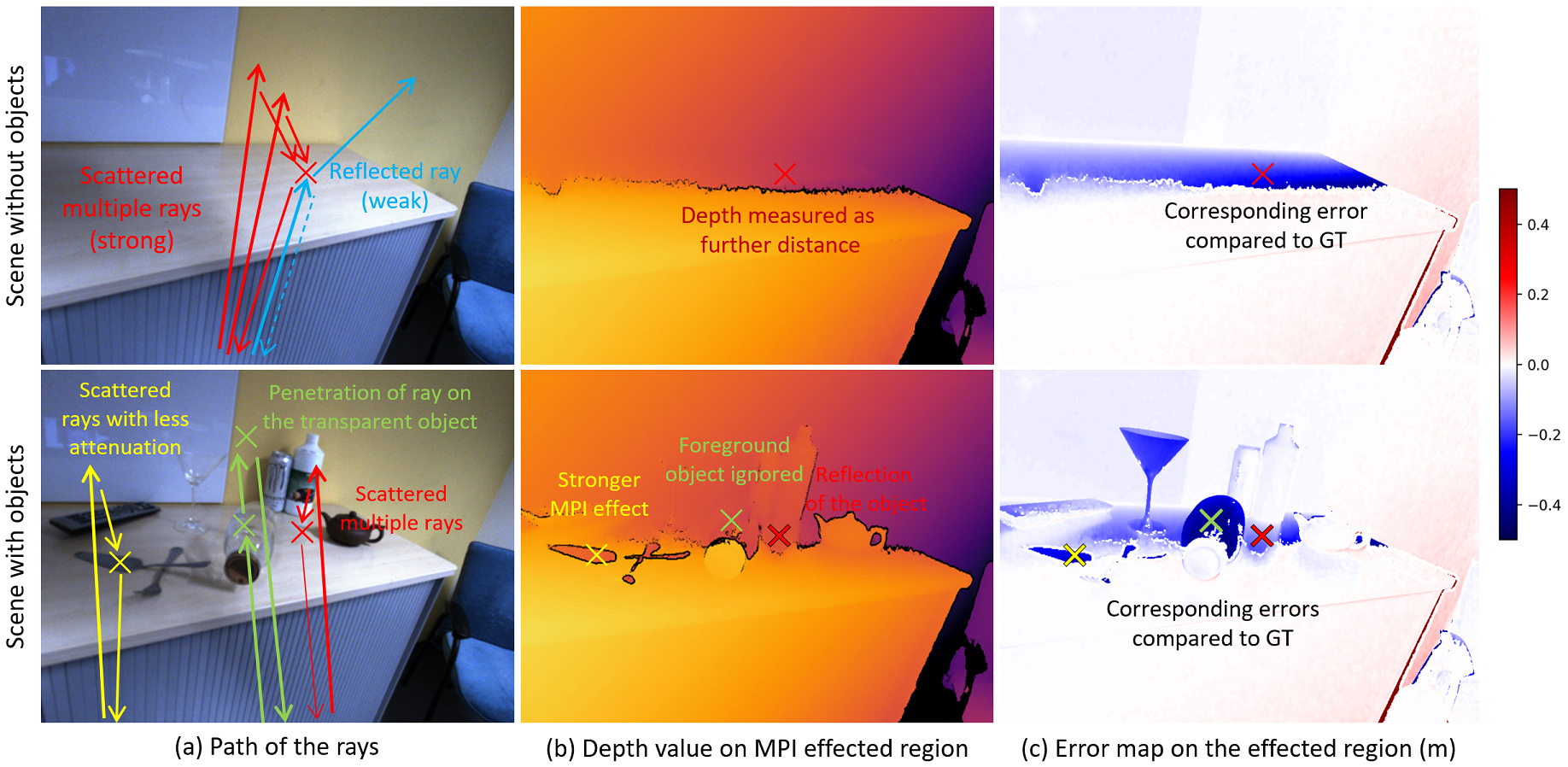}
    \caption{Detailed ray paths with MPI and surface material induced error on D-ToF modality. While D-ToF produces dense and sharp depth, its quality is highly dependent on the surface material and the incident angle.}
    \label{fig:dtof_mpi}
\end{figure*}

\begin{figure*}[!htbp]
 \centering
    \includegraphics[width=\linewidth]{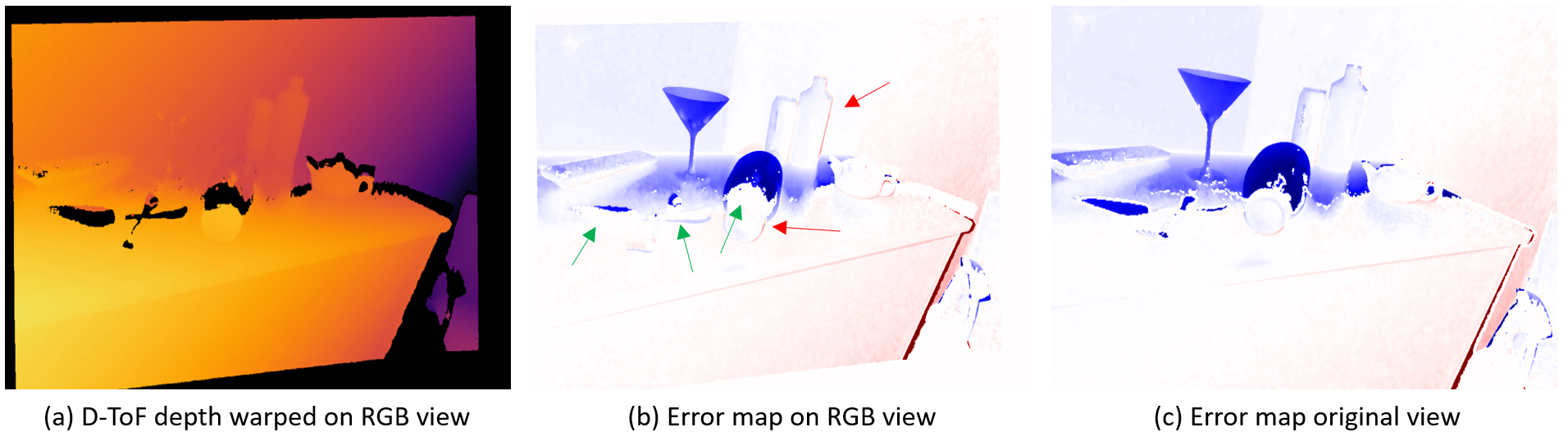}
    \caption{Error after warping D-ToF into RGB view. Slight errors are introduced on some edges (red) while expansion of the invalid area helps to invalidate on the reflective objects (green).}
    \label{fig:dtof_aligned}
\end{figure*}

As mentioned in Subsec.~\ref{subsec:dtof_camera}, D-ToF modality suffers from its own reflection-based nature, such as MPI and material dependent artefacts. When the angle of the surface normal of the scene is close to the incident angle of the infrared signal, the strength of the reflected signal becomes weak due to scattering effects (Fig.~\ref{fig:dtof_mpi}, (a) blue arrow) while multiple scattered signals from the other surfaces which has more traveling distance are received and with stronger strength (Fig.~\ref{fig:dtof_mpi}, (a) red arrow) and interfere with the original signal (MPI), producing a wrong measurement of the depth on the area with further distance which looks like a reflection or shadow of the object to the surface (Fig.~\ref{fig:dtof_mpi}, (b) red marker). This effect can be intensified when the surface material is reflective, which gives even stronger artefact as its reflective surface bounces even weaker and noisier signal with less attenuation (Fig.~\ref{fig:dtof_mpi}, (a,b) yellow arrow\&marker). On the other hands, when the surface material is transparent, the emitted infrared signal rather goes through the object in the both ways  (Fig.~\ref{fig:dtof_mpi}, (a) green arrow) which at the end ignores the object and the sensor produce the depth value as similar level as its background (Fig.~\ref{fig:dtof_mpi}, (b) green marker - material dependent artefact). Quality of the depth map degrades slightly around some boundaries after warping into the RGB frame (Fig.~\ref{fig:dtof_aligned}, (b), red), while the invalid regions actually helps to invalidate more area on wrong depth especially on the reflective objects  (Fig.~\ref{fig:dtof_aligned}, (b), green) , which might become beneficial when it is used in the training.

\subsubsection{I-ToF Camera}
\label{subsec:itof_camera_error}
\begin{figure*}[!tbp]
 \centering
    \includegraphics[width=\linewidth]{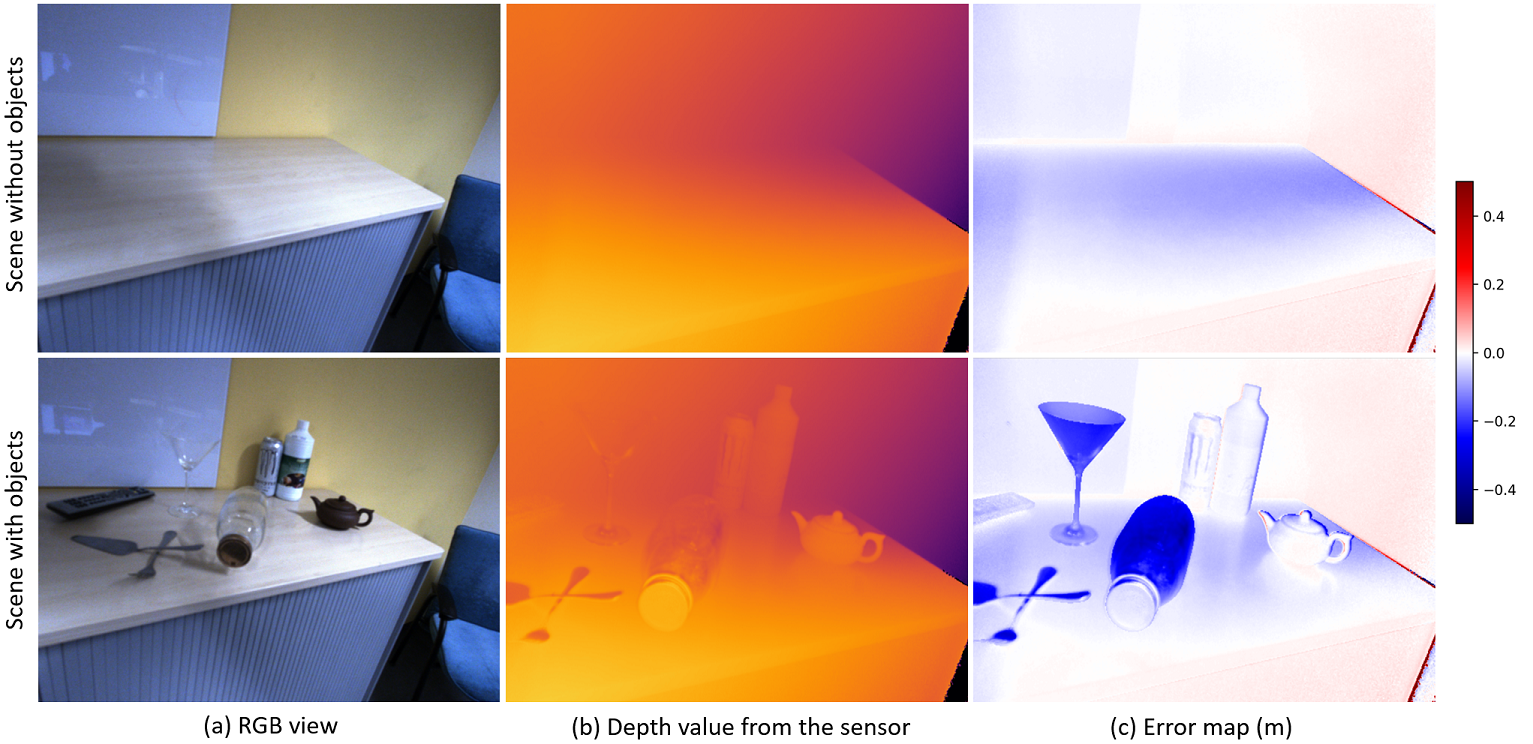}
    \caption{Depth quality from I-ToF camera. I-ToF modality suffers from same type of artefect as D-ToF. While depth map itself is more sense and suffers less from MPI artefact on the table.}
    \label{fig:itof_mpi}
\end{figure*}

\begin{figure*}[!tbp]
 \centering
    \includegraphics[width=\linewidth]{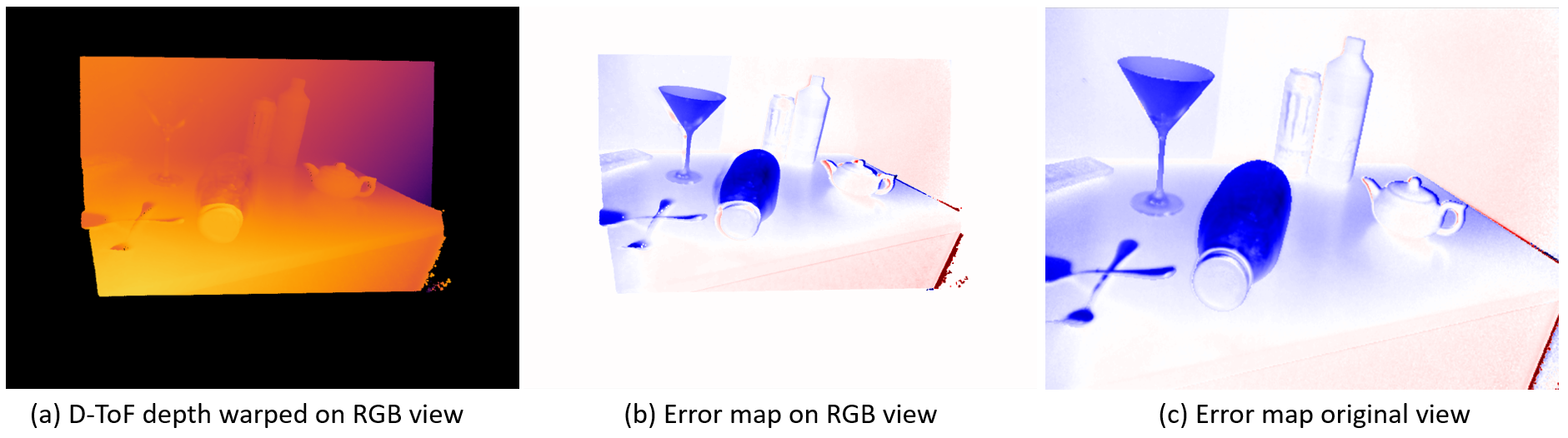}
    \caption{Error after warping I-ToF into RGB view. Not like D-ToF, most of depth error exists without being invalidated, which might introduce more error when it used as GT during the training.}
    \label{fig:itof_aligned}
\end{figure*}

As mentioned in Subsec.~\ref{subsec:itof_camera}, I-ToF modality suffers by its own reflection based nature as well similar to D-ToF, such as MPI and material dependent artefact (Fig.~\ref{fig:itof_mpi}. Although the quality of depth itself seems better as the depth itself is more dense (with less invalid region) and amount of the artefacts are less, it is hard to say I-ToF modality is better than D-ToF as these two camera are in different price range and power level. Also less invalid area but rather with wrong depth didn't help invalidating depth (Fig.~\ref{fig:itof_aligned}) not like in D-ToF case, which could result in artefact in the prediction when it is used as GT during the training.

\subsubsection{Active Stereo Camera}
\label{subsec:active_stereo_camera_error}
\begin{figure*}[!htbp]
 \centering
    \includegraphics[width=\linewidth]{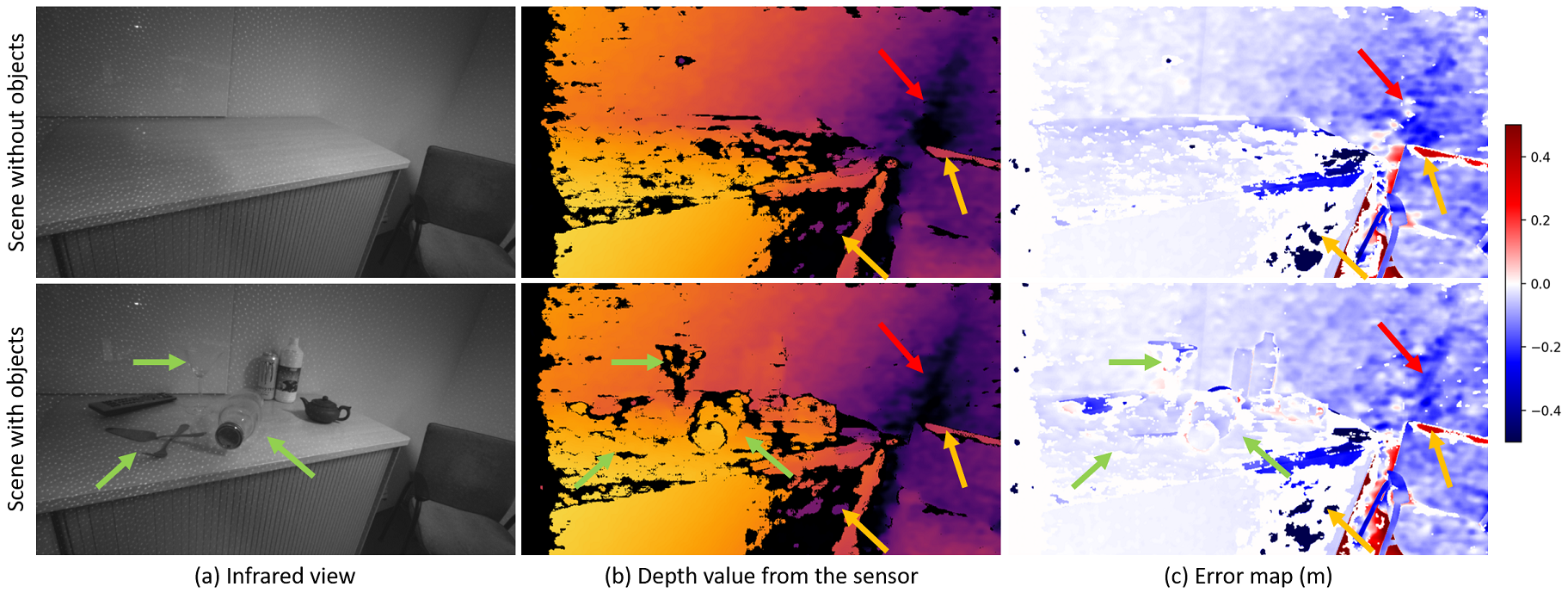}
    \caption{Depth quality from Active Stereo camera. While depth map suffers less on the challenging material, quality of depth itself is far behind either of ToF modality in multiple aspects, such as sharpness, variance, sparsity.}
    \label{fig:d435_not_aligned}
\end{figure*}

\begin{figure*}[!htbp]
 \centering
    \includegraphics[width=\linewidth]{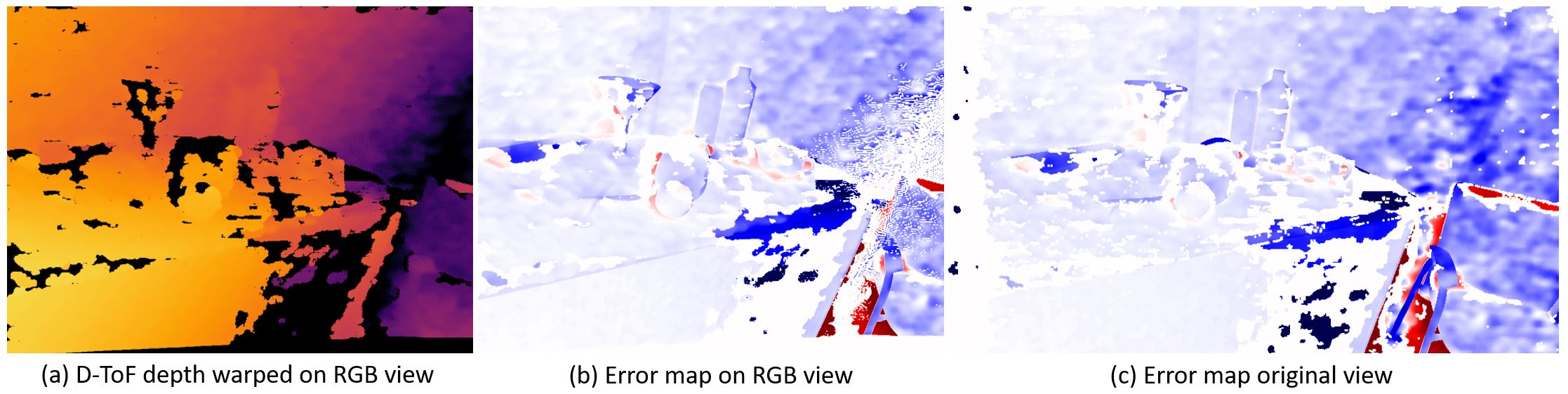}
    \caption{Error after warping Active Stereo into RGB view. Note that there isn't significant change in the depth quality after the warping.}
    \label{fig:d435_aligned}
\end{figure*}

As the stereo camera uses left and right matching with photoelectric cue, depth map suffers less on the challenging material as the projection can be visible on the surface as well as left-right check can be performed to invalidate region with the wrong depth. For this reason, depth on glass or the reflective object is significantly more accurate compared to either of ToF modality (Fig.~\ref{fig:d435_not_aligned}, green arrow). On the other hands, due to its nature of pattern projection far distance that depth quality gets worsen as the scene gets further (Fig.~\ref{fig:d435_not_aligned}, red arrow) the projection pattern gets attenuated and spread in the far distance. Moreover, the depth map in general is more blurry, jittery, sparse and has wrong values on some regions without being invalidated (Fig.~\ref{fig:d435_not_aligned}, orange arrow) which can introduce negative influence when it is used as GT, such as blurriness and depth jittering. Error introduced by warping is trivial (Fig.~\ref{fig:d435_aligned}) as the original depth map is already blurry and sparse.

\clearpage
\newpage

\subsection{Detailed Background and Objects Description}
\label{sec:bckgr_obj_description}

\begin{figure*}[!b]
 \centering
    \includegraphics[width=\linewidth]{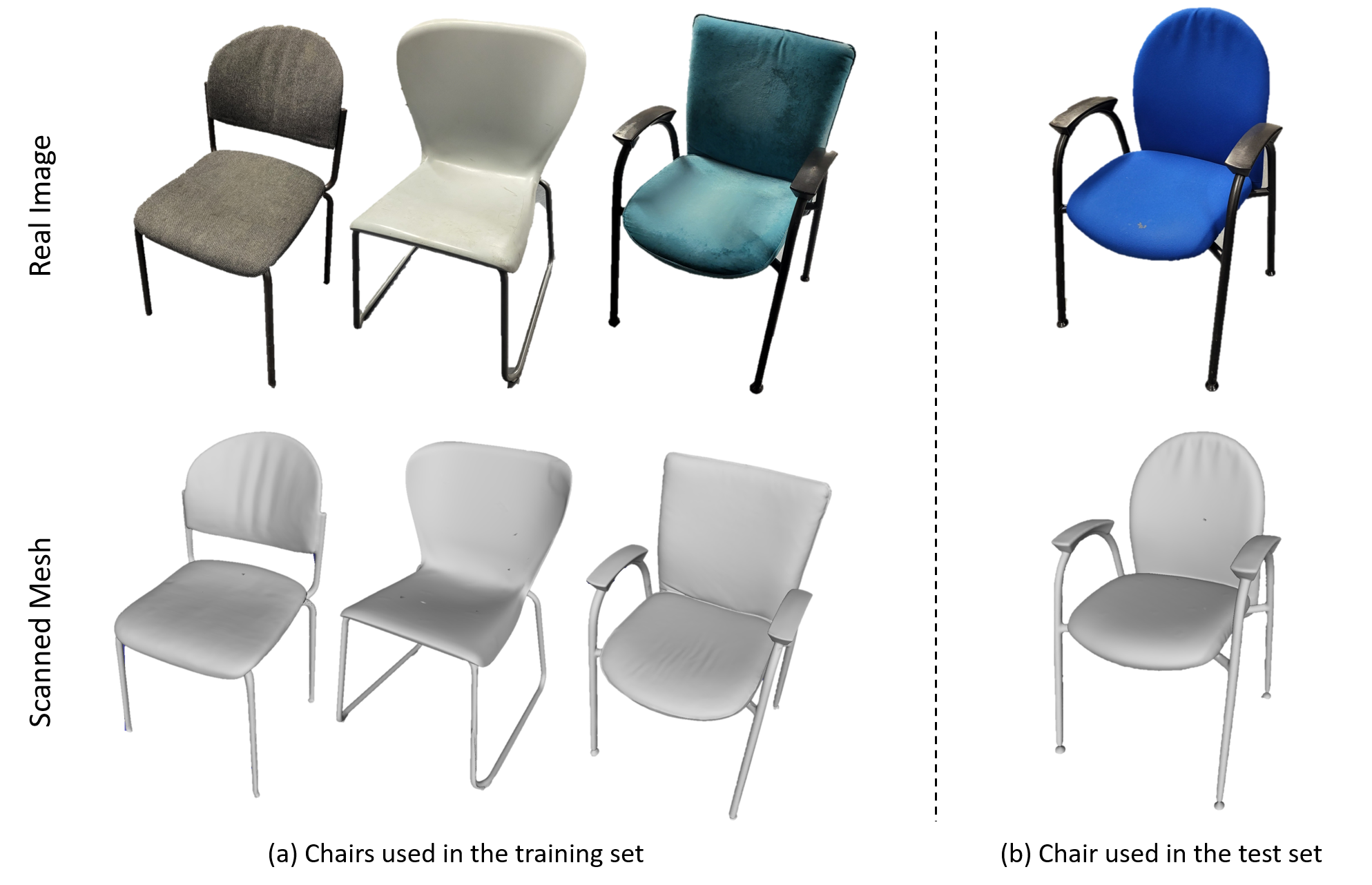}
    \caption{Chairs used in the dataset. Chairs in group (a) are used for the training set and the chair in (b) is used for the test set.}
    \label{fig:chair}
\end{figure*} 

\begin{figure*}[!p]
 \centering
    \includegraphics[width=\linewidth]{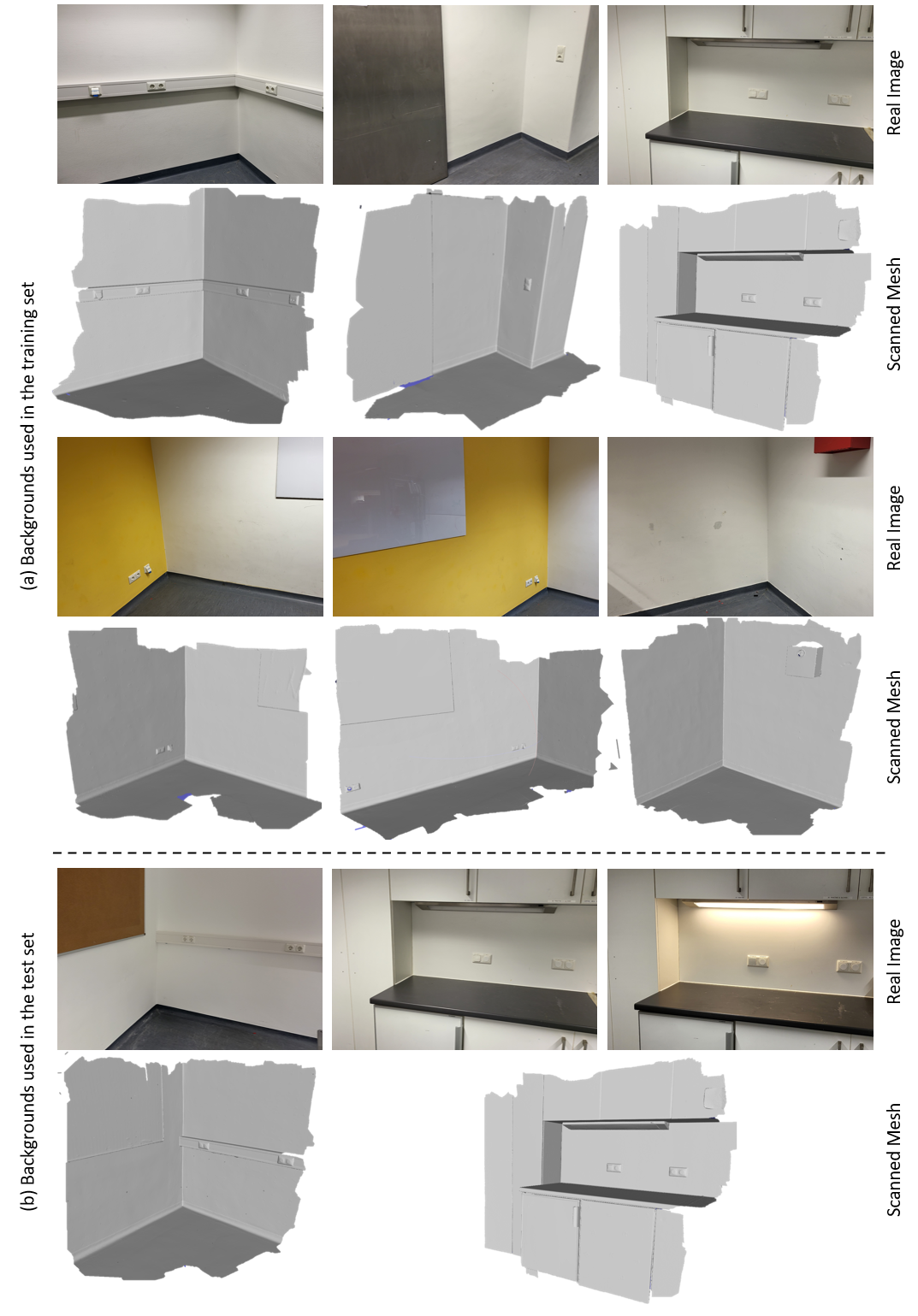}
    \caption{Backgrounds used in the dataset. Note that one of the background in the group (b) is also included in the training set, but we varied the lighting condition to provide different various factors for evaluation.}
    \label{fig:wall}
\end{figure*}

\begin{figure*}[!t]
 \centering
    \includegraphics[width=\linewidth]{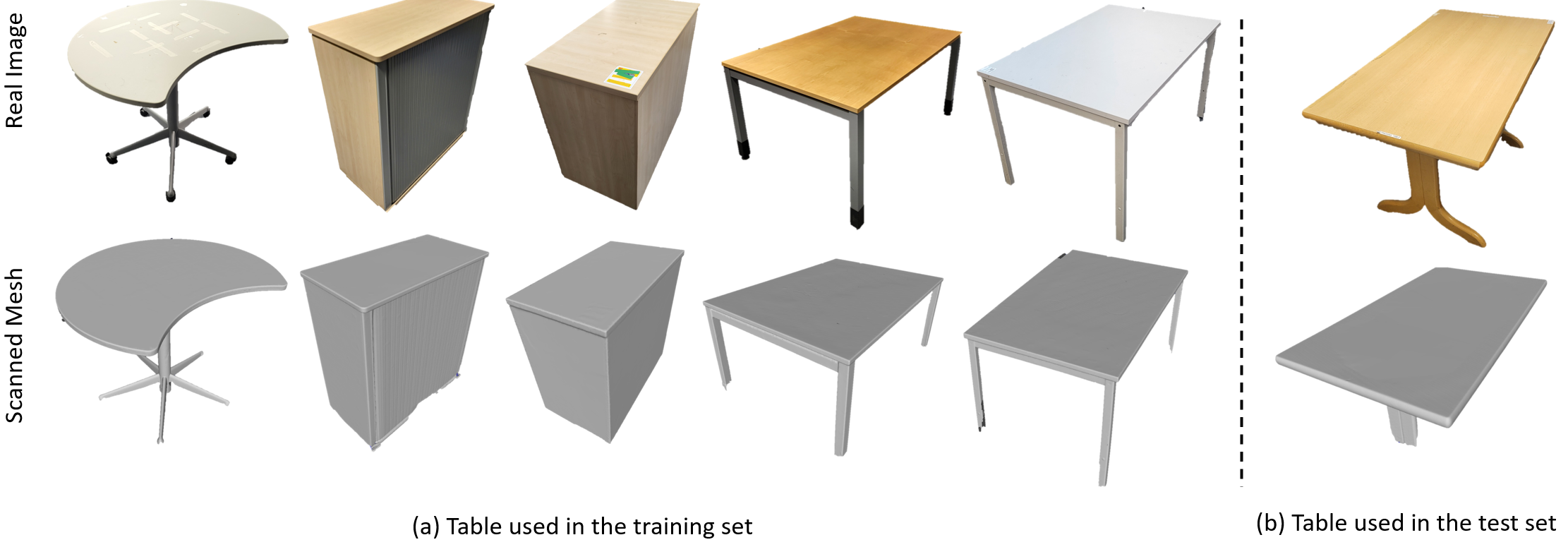}
    \caption{Tables used in the dataset. Tables in group (a) are used for the training set and the table in (b) is used for the test set. Note that, unlike small objects or chairs, we decide not to scan some parts of the large tables (e.g. end of their legs) as the cameras cannot see the part in their trajectories.}
    \label{fig:table}
\end{figure*}

\begin{figure*}[!b]
 \centering
    \includegraphics[width=\linewidth]{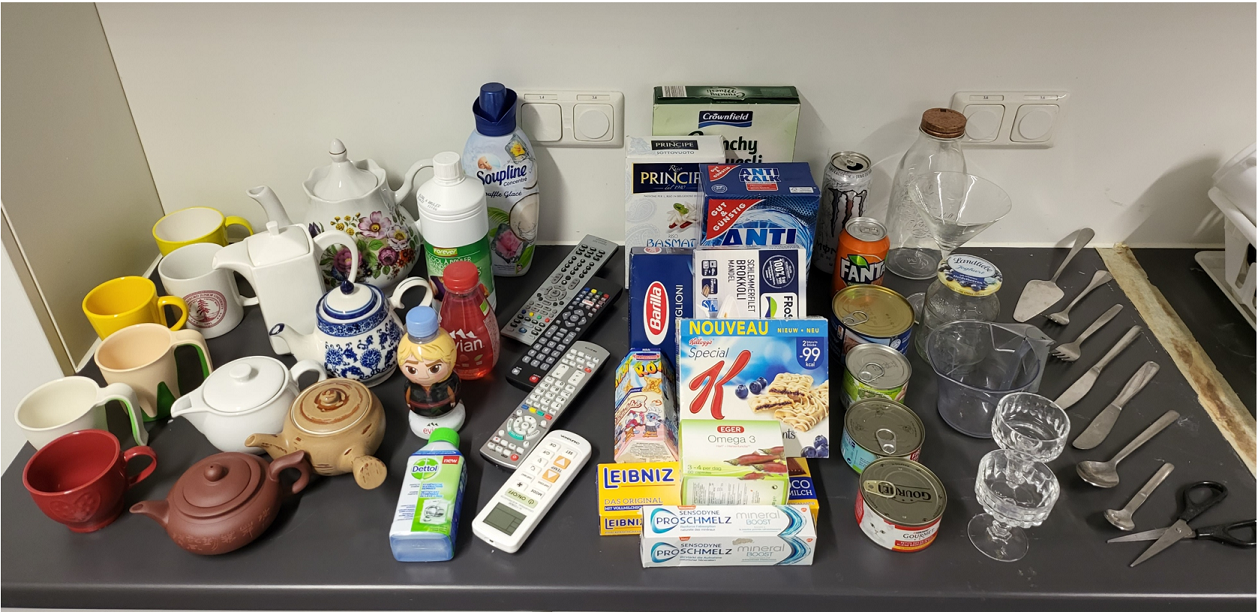}
    \caption{Collection of small household objects used in the training set. Objects from 8 household categories are used in the training set, 3 of which have photometrically challenging surface material - partially reflective (can), transparent (glass/plastic), reflective (cutlery).}
    \label{fig:obj_train}
\end{figure*}

\begin{figure*}[!t]
 \centering
    \includegraphics[width=\linewidth]{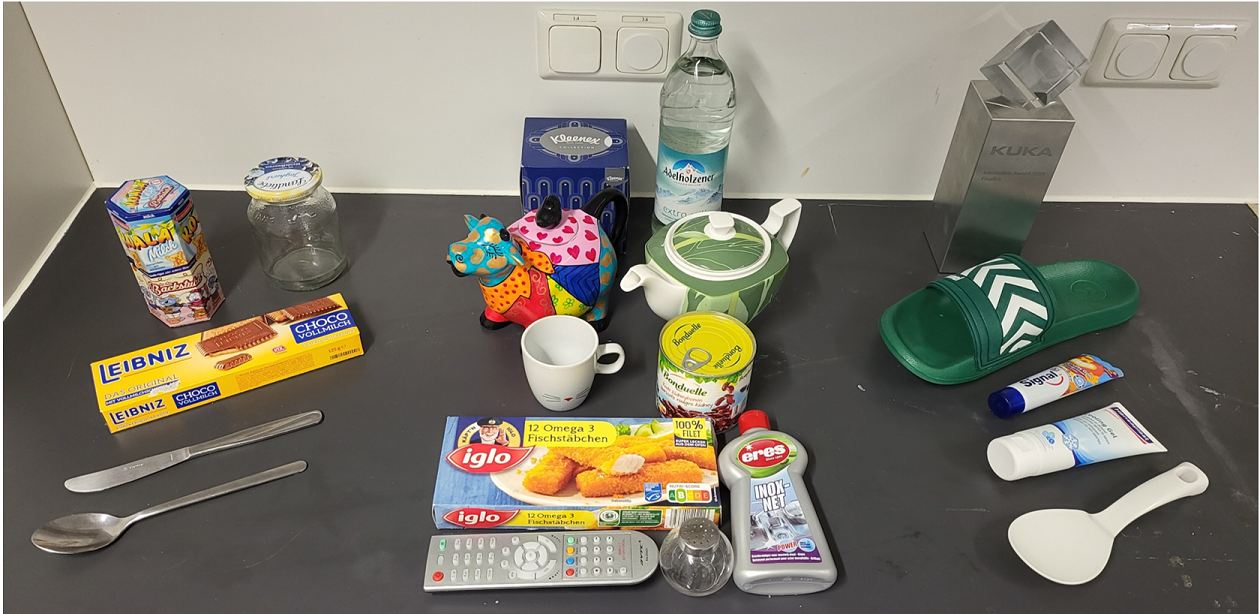}
    \caption{Collection of small household objects used in the test set. The test set comprises a mixture of seen (left column) and unseen (mid column) objects from 8 seen categories and a few objects from unseen categories (right column - tube, slipper, plastic kitchenware, trophy) are used.}
    \label{fig:obj_test}
\end{figure*}

As described in Sec.~4 in the main paper, our dataset comprises a total of 13 scenes divided into 10 scenes for training and 3 testing scenes composed of a mixture of 4 different chairs, 6 different tables, 64 household objects from 8 plus 4 different categories (i.e. cup, teapot, bottle, remote, boxes, can, glass, cutlery and tube, shoe, plastic kitchenware, trophy) and and 7 different indoor areas. Test sets have 1 unseen background and 2 seen backgrounds with and without different lighting and contain a mixture of seen/unseen objects from seen/unseen categories. In this section, we show detailed images of backgrounds, chairs, tables, and other objects. Fig.~\ref{fig:chair} and~\ref{fig:table} respectively show images of 3 chairs and 6 tables used in the dataset and their corresponding meshes. Fig.~\ref{fig:obj_train} and~\ref{fig:obj_test} show a collection of household objects used in training and test set. Fig.~\ref{fig:wall} shows 9 backgrounds used in the dataset and their corresponding meshes.

\subsubsection{Detailed Scene Description}
\label{sec:scene_description}

\begin{figure*}[!htbp]
 \centering
    \includegraphics[width=\linewidth]{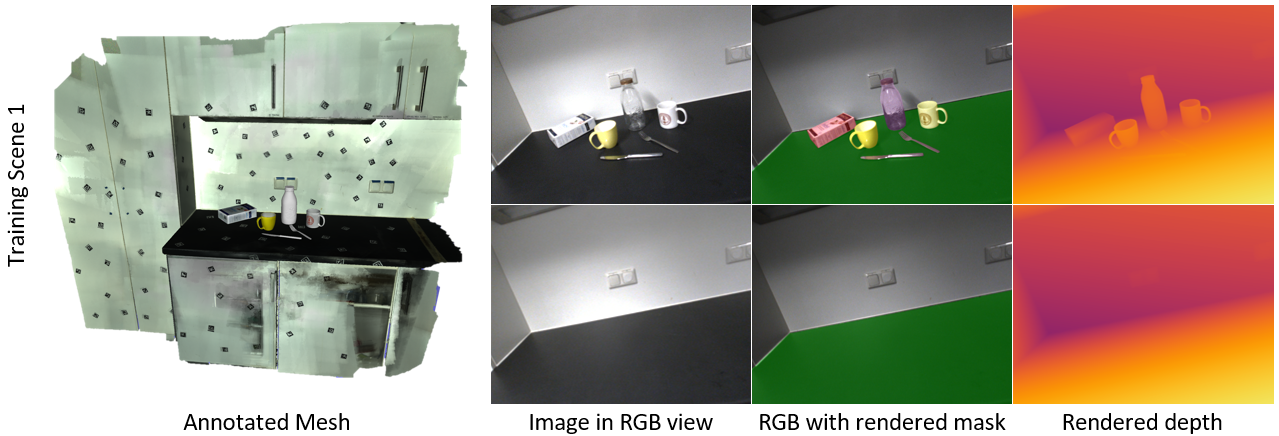}
    \caption{Example images from Training Scene 1. The annotated mesh is shown on the left together with an RGB view from the scene (second from left) with and without objects. The overlayed masks (second from right) and the rendered depth (right) illustrate the annotation quality of our data.}
    \label{fig:wall1}
\end{figure*}

\begin{figure*}[!htbp]
 \centering
    \includegraphics[width=\linewidth]{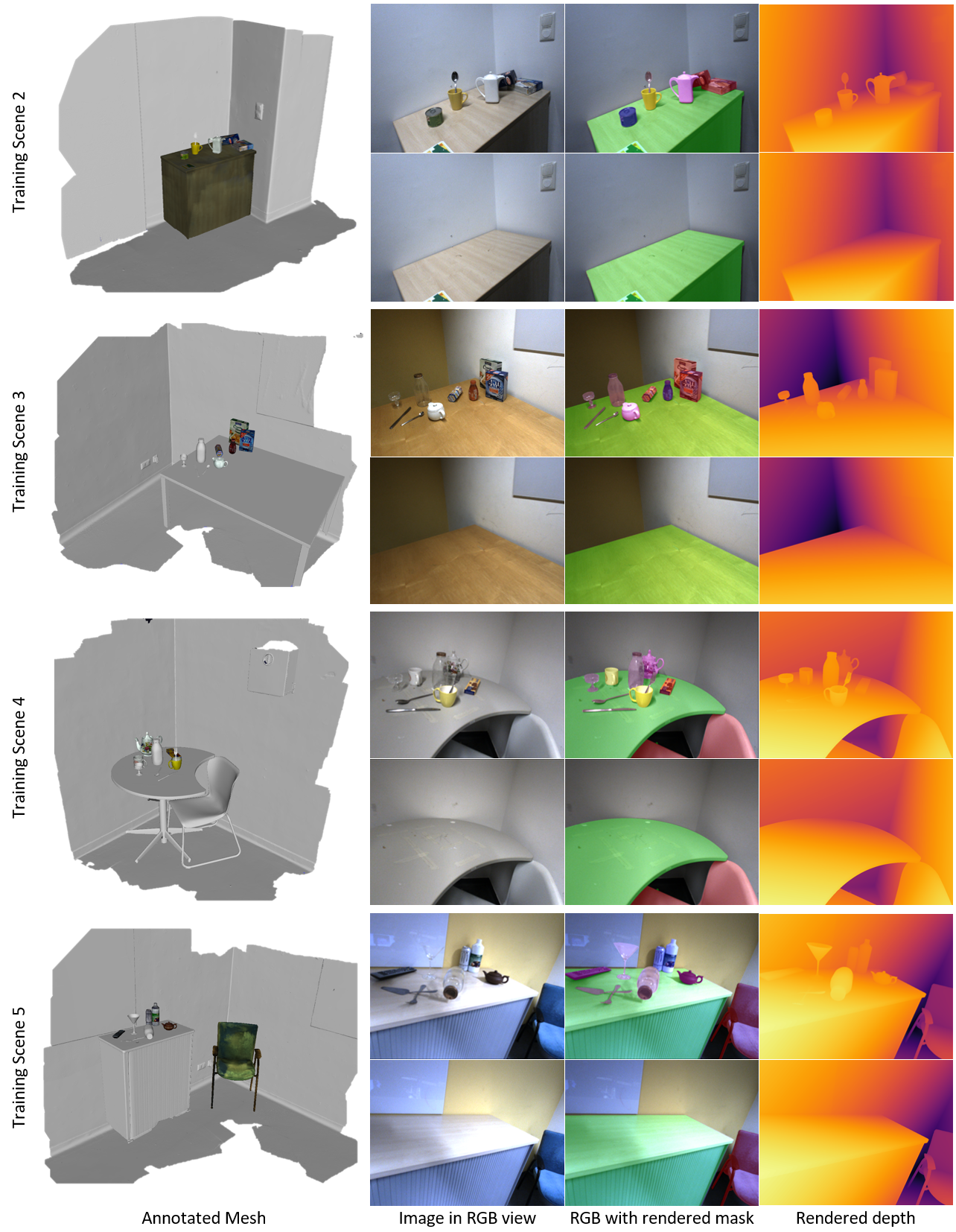}
    \caption{Example images from Training Scene 2-5. The annotated mesh for 4 different scenes is shown on the left together with an RGB view from the scene (second from left) with and without objects. The overlayed masks (second from right) and the rendered depth (right) illustrate the annotation quality of our data.}
    \label{fig:wall2}
\end{figure*}

\begin{figure*}[!htbp]
 \centering
    \includegraphics[width=\linewidth]{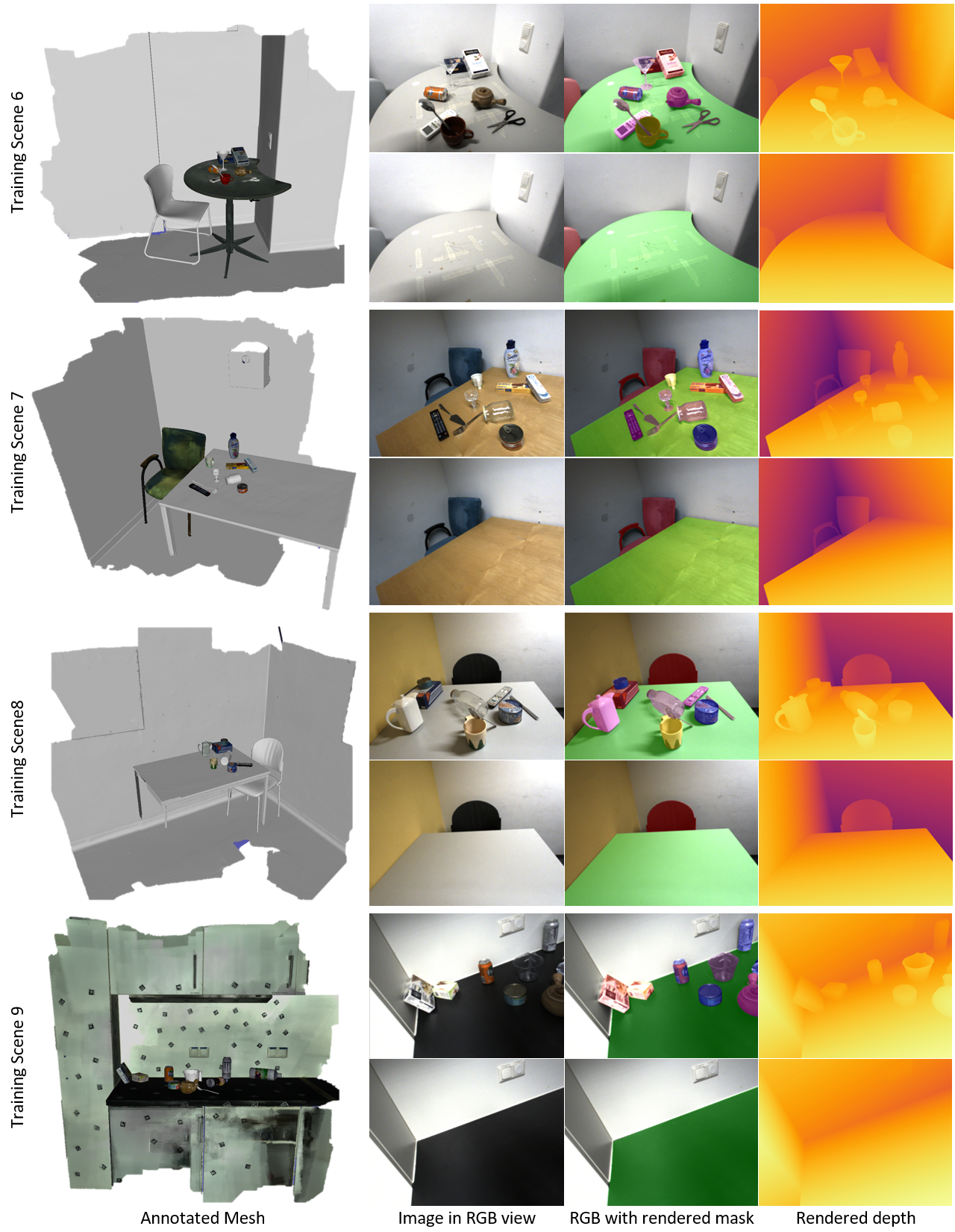}
    \caption{Example images from Training Scene 6-9. The annotated mesh for four different scenes is shown on the left together with an RGB view from the scene (second from left) with and without objects. The overlayed masks (second from right) and the rendered depth (right) illustrate the annotation quality of our data.}
    \label{fig:wall3}
\end{figure*}

\begin{figure*}[!htbp]
 \centering
    \includegraphics[width=\linewidth]{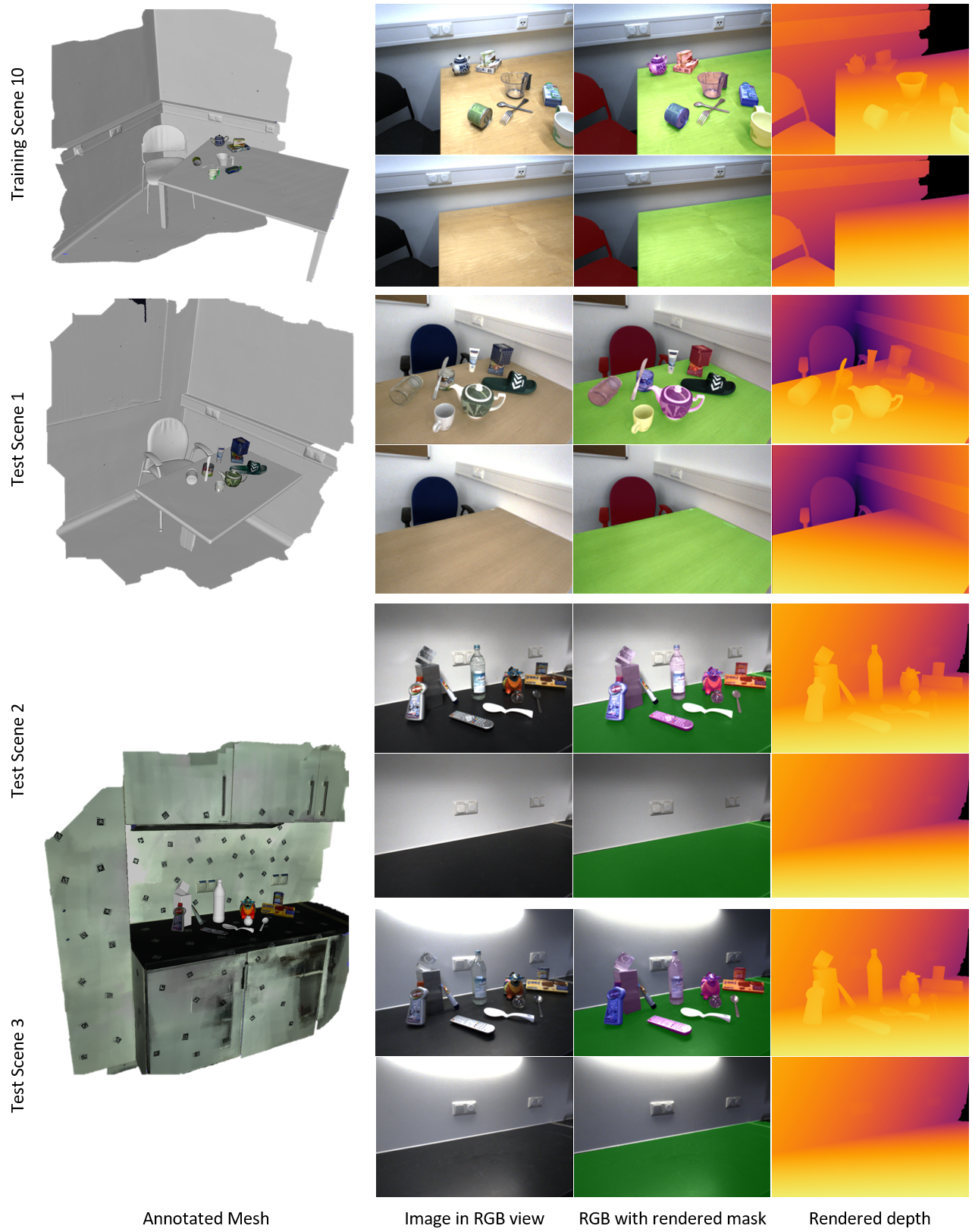}
    \caption{Example images from Training Scene 10 and Test scene 1-3. The annotated mesh is shown on the left together with an RGB view from the scene (second from left) with and without objects. The overlayed masks (second from right) and the rendered depth (right) illustrate the annotation quality of our data. Note that the test scene 2,3 are recorded in the exactly same pose and trajectory but with the different lighting.}
    \label{fig:wall4}
\end{figure*}

As described, our training set is composed of 10 scenes, and the test set is composed of 3 scenes. For each scene, we include 2 different trajectories. Each trajectory covers 2 setups with and without objects (naked scene). This sums up to 800-1200 frames per scene and a total of ca.~10k frames. In this section, we show several sample images of the scenes in Fig.~\ref{fig:wall1},~\ref{fig:wall2}, and~\ref{fig:wall3}, ~\ref{fig:wall4}. Each of them consists of an annotated mesh and RGB images with different types of rendering, which show the diversity and quality of our dataset.

\subsubsection{Partial Scanning of the Scene and Mesh Fitting}
\label{sec:partial_scanning}

\begin{figure*}[!htbp]
 \centering
    \includegraphics[width=\linewidth]{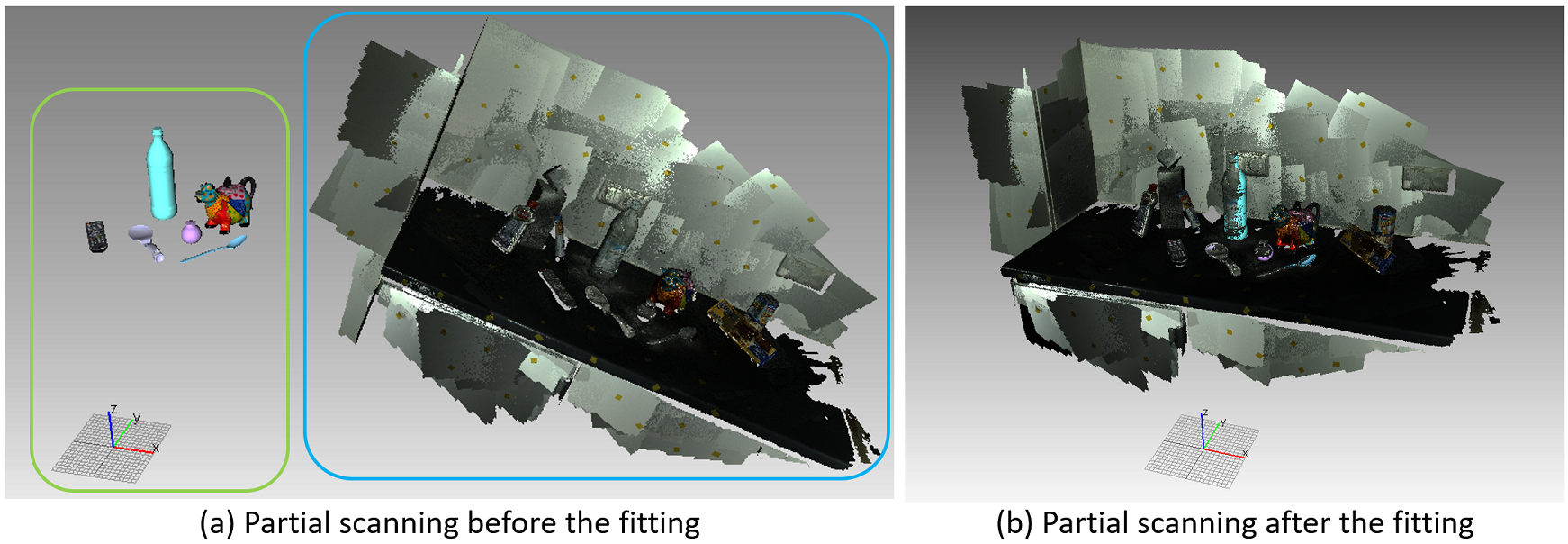}
    \caption{Example of partial scanning of the scene before and after the fitting on scene 13. Note that the center of the partial scanned mesh is aligned to robot base (xyz coordinate marker) after fitting it onto the mesh of the annotated objects.}
    \label{fig:annotated_objects}
\end{figure*}

\begin{figure*}[!htbp]
 \centering
    \includegraphics[width=\linewidth]{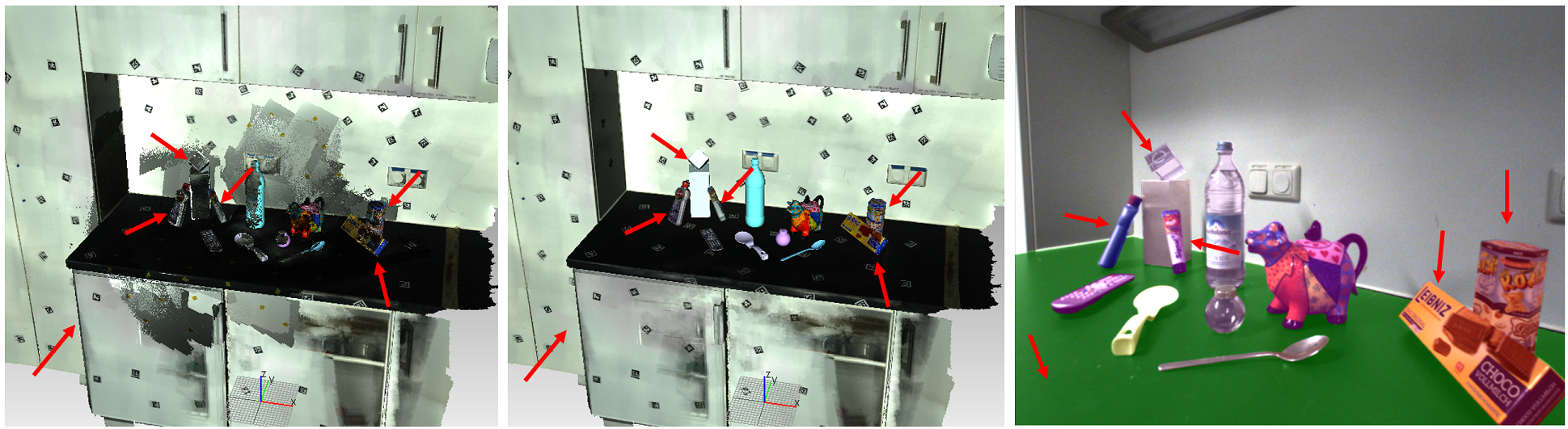}
    \caption{Example of far objects and background fitting onto partially scanned mesh. Left: Background and objects are fit to partial scans. Centre: All annotated meshes are shown without partial scans. Right: Corresponding scene from the camera viewpoint with augmented object masks. Note that the annotation quality of meshes with partial scans and robot arm is similar. The annotated meshes via partial scanning are marked with red arrows.}
    \label{fig:fitting_final}
\end{figure*}

As mentioned in Sec.~3 in the main paper, we use partial scanning and mesh fitting to annotate background, large objects, and objects outside the robotic workspace. This section shows images of partial scanning and the mesh fitting from one of the scenes as an example. The green box in Fig.~\ref{fig:annotated_objects}, (a) shows annotated meshes of the objects by the robotic arm. Once the objects are annotated, the scene is partially scanned with multiple viewpoints to make the scanning dense and cover multiple facets of the background. Note that the center of the scanning is not yet in the robot base coordinates (Fig.~\ref{fig:annotated_objects}, (a) blue box). Once the partial scanning is done, the scanned mesh is then fit onto the annotated objects, such that the partially scanned mesh origin concides with the robot base (Fig.~\ref{fig:annotated_objects}, (b)). Once the scanned mesh is put to robot base coordinates, we fit background, large objects, and distant objects meshes also in robot base coordinates to annotate them (Fig.~\ref{fig:fitting_final}, (a)). Fig.~\ref{fig:fitting_final}, (b-c) shows the result of the annotated mesh. %As this method may contain errors propagated from noise and error of the mesh, we manually refined the pose of the fitted mesh in few mm in translation and sub degree in rotation such that the objects can be well globally well aligned into all of the cameras' viewpoints. 
For the mesh fitting, we used Artec Studio 10 Professional (Artec 3D, Luxembourg) which runs a point correspondence and ICP-based method to fit the meshes.

\clearpage
\newpage
%%%%%%%%% REFERENCES
{\small
\bibliographystyle{ieee_fullname}
\bibliography{PaperForReview_supp}
}